\documentclass[10pt,journal,cspaper,compsoc]{IEEEtran}

\usepackage{booktabs} % Allows the use of \toprule, \midrule and
\usepackage{epsfig}
\usepackage{graphicx}
\usepackage{amsmath}
\usepackage{amssymb}
\usepackage{algorithm}
\usepackage{algorithmic}
\usepackage{subfigure}
\usepackage{url}
\usepackage{color}
\usepackage{multirow}
\usepackage[usenames,dvipsnames,table]{xcolor}
\usepackage{times}

\usepackage[pagebackref=true,breaklinks=true,letterpaper=true,colorlinks,bookmarks=false]{hyperref}

\begin{document}
	
	%%%%%%%%% TITLE
	\title{Proposal-free Network for Instance-level Object Segmentation}
	
	\author{Xiaodan~Liang, Yunchao~Wei, Xiaohui~Shen, Jianchao~Yang, Liang~Lin, Shuicheng~Yan~\IEEEmembership{Senior Member,~IEEE}
		% <-this % stops a space
		\IEEEcompsocitemizethanks{\IEEEcompsocthanksitem Xiaodan Liang is with Department of Electrical and Computer Engineering, National University of Singapore, and also with Sun Yat-sen University. Yunchao Wei is with Department of Electrical and Computer Engineering, National University of Singapore, and also with the Institute of Information Science, Beijing Jiaotong University. Xiaohui Shen is with Adobe Research. Jianchao Yang is with Snapchat Research. Liang Lin is with Sun Yat-sen Unviersity. Shuicheng Yan is with Department of Electrical and Computer Engineering, National University of Singapore.}\protect\\
		
		% <-this % stops a space
		\thanks{}}
	
	\markboth{IEEE TRANSACTIONS ON PATTERN ANALYSIS AND MACHINE INTELLIGENCE, VOL. XX, NO. X, X 20XX}% MARCH 2012
	{Shell \MakeLowercase{\textit{et al.}}: Bare Demo of IEEEtran.cls for Computer Society Journals}

	
	%\thispagestyle{empty}
	
	%%%%%%%%% ABSTRACT
	\IEEEcompsoctitleabstractindextext{%
		\begin{abstract}
			Instance-level object segmentation is an important yet under-explored task. The few existing studies are almost all based on region proposal methods to extract candidate segments and then utilize object classification to produce final results. Nonetheless, generating accurate region proposals itself is quite challenging. In this work, we propose a Proposal-Free Network (PFN ) to address the instance-level object segmentation problem, which outputs the instance numbers of different categories and the pixel-level information on 1) the coordinates of the instance bounding box each pixel belongs to, and 2) the confidences of different categories for each pixel, based on pixel-to-pixel deep convolutional neural network. All the outputs together, by using any off-the-shelf clustering method for simple post-processing, can naturally generate the ultimate instance-level object segmentation results. The whole PFN can be easily trained in an end-to-end way without the requirement of a proposal generation stage. Extensive evaluations on the challenging PASCAL VOC 2012 semantic segmentation benchmark demonstrate that the proposed PFN solution well beats the state-of-the-arts for instance-level object segmentation. In particular, the $AP^r$ over 20 classes at 0.5 IoU reaches 58.7\% by PFN, significantly higher than 43.8\% and 46.3\% by the state-of-the-art algorithms, SDS~\cite{hariharan2014simultaneous} and~\cite{liu2015multi}, respectively.

		\end{abstract}
	}
	
	\maketitle
	
	\IEEEdisplaynotcompsoctitleabstractindextext
	
	\IEEEpeerreviewmaketitle

%%%%%%%%% BODY TEXT

\section{Introduction}
\vspace{-2mm}

Over the past few decades, two of the most popular object recognition tasks, object detection and semantic segmentation, have received a lot of attention. The goal of object detection is to accurately predict the semantic category and the bounding box location for each object instance, which is a quite coarse localization. Different from object detection, the semantic segmentation task aims to assign the pixel-wise labels for each image but provides no indication of the object instances, such as the  object instance number and precise semantic region for any particular instance. In this work, we follow some of the recent works~\cite{hariharan2014simultaneous}~\cite{liu2015multi}~\cite{zhang2015monocular} and attempt to solve a more challenging task, instance-level object segmentation, which predicts the segmentation mask for each instance of each category. We suggest that the next generation of object recognition should provide a richer and more detailed parsing for each image by labeling each object instance with an accurate pixel-wise segmentation mask. This is particularly important for real-world applications such as image captioning, image retrieval, 3-D navigation and driver assistance, where describing a scene with detailed individual instance regions is potentially more informative than describing roughly with located object detections. However, instance-level object segmentation is very challenging due to high occlusion, diverse shape deformation and appearance patterns, obscured boundaries with respect to other instances and background clutters in real-world scenes. In addition, the exact instance number of each category within an image is dramatically different.

\begin{figure*}
	\begin{center}
		\includegraphics[scale=0.13]{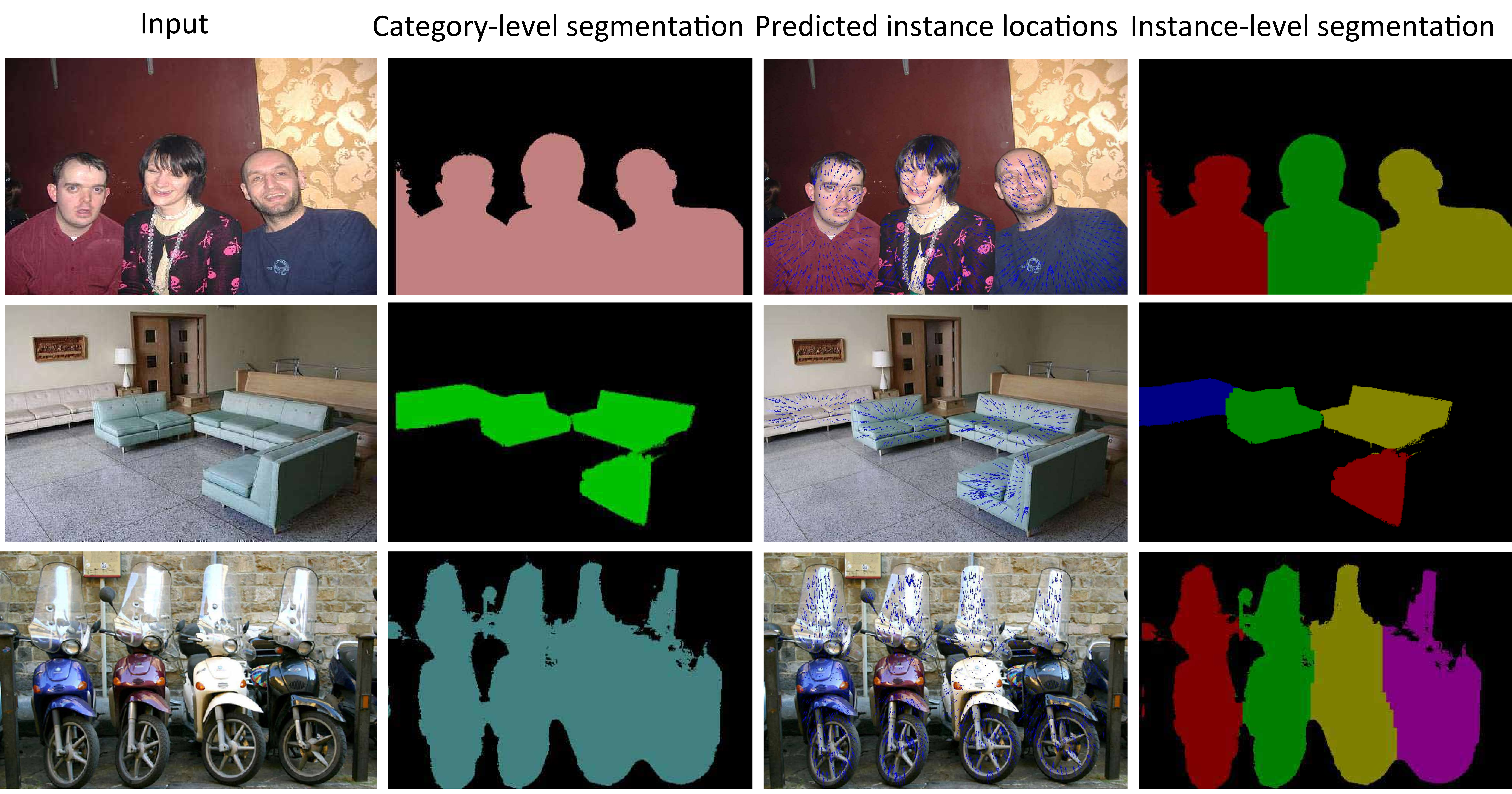}
		\vspace{-2mm}
		\caption{{Exemplar instance-level object segmentation results. For each image, the category-level segmentation results, predicted instance locations for all foreground pixels and instance-level segmentation results are sequentially shown in each row. Different colors indicate the different object instances for each category. To better show the predicted instance locations, we plot velocity vectors that start from each pixel to its corresponding predicted instance center as shown by the arrow. Note that the pixels predicting similar object centers can be directly collected as one instance region. Best view in color and {scale up three times.}}}
		\vspace{-8mm}
		\label{fig:motivation}
	\end{center}
\end{figure*}

Recently, tremendous advances in semantic segmentation~\cite{BoxSup}~\cite{lin2015efficient}~\cite{long2014fully}~\cite{CRF-RNN} and object detection~\cite{fastercnn}~\cite{redmon2015you}~\cite{stewart2015end} have been made relying on deep convolutional neural networks (DCNN)~\cite{szegedy2014going}~\cite{vgg}. Some previous works have been proposed to address instance-level object segmentation. Unfortunately, none of them has achieved excellent performance in an end-to-end way. In general, these previous methods take complicated pre-processing such as bottom-up region proposal generation~\cite{MCG}~\cite{uijlings2013selective}~\cite{zitnick2014edge}~\cite{pinheiro2015learning} or post-processing such as graphical inference as the requisite. Specifically, the recent two approaches, SDS~\cite{hariharan2014simultaneous} and the one proposed by Chen et al.~\cite{liu2015multi}, use the region proposal methods to first generate potential region proposals and then classify on these regions. After classification, post-processing such as non-maximum suppression (NMS) or Graph-cut inference, is used to refine the regions, eliminate duplicates and rescore these regions. Note that most region proposal techniques~\cite{MCG}~\cite{uijlings2013selective}~\cite{pinheiro2015learning} typically generate thousands of  potential regions, and take more than one second per image. Additionally, these proposal-based approaches often fail in the presence of strong occlusions. When only small regions are observed and evaluated without awareness of the global context, even a highly accurate classifier can produce many false alarms. Depending on the region proposal techniques, the common pipelines are often trained using several independent stages. These separate pipelines rely on independent techniques at each stage and the targets of the stages are significantly different. For example, the region proposal methods try to maximize region recalls while the classification optimizes for single class accuracy.

\begin{figure*}
	\begin{center}
		\includegraphics[scale=0.55]{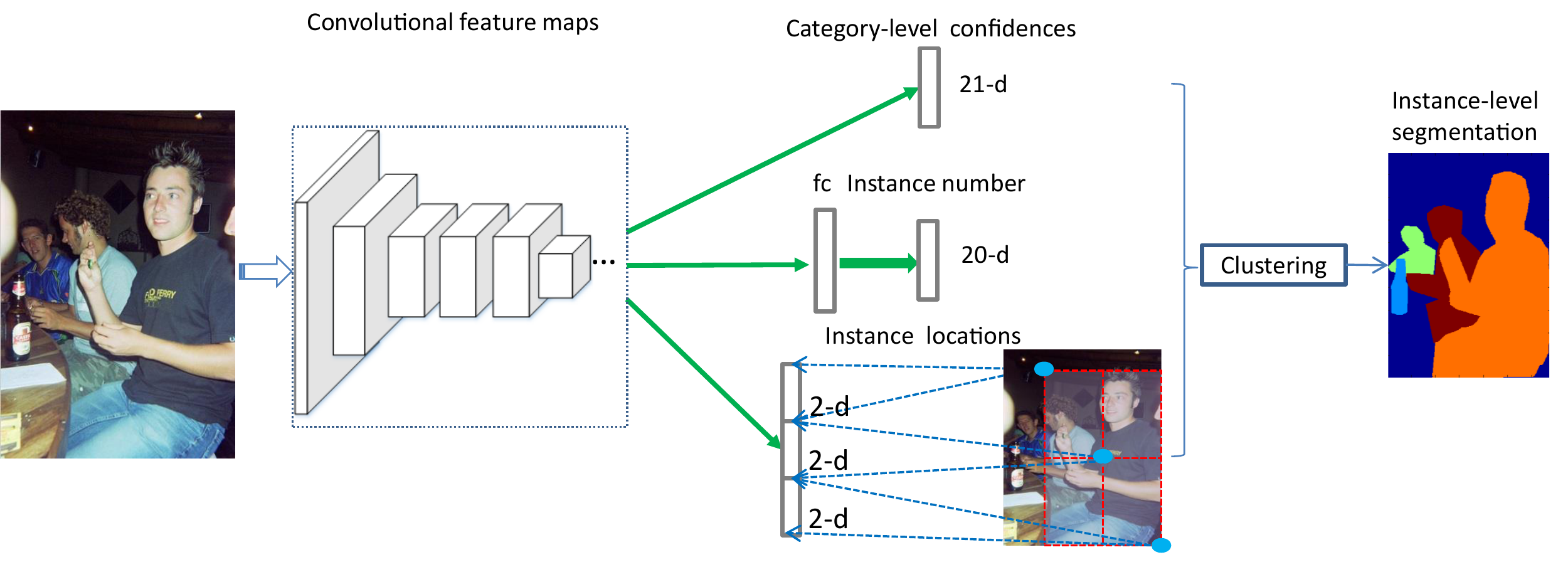}
		\vspace{-4mm}
		\caption{{The proposal-free network overview. Our network predicts the instance numbers of all categories and the pixel-level information that includes the category-level confidences for each pixel and the coordinates of the instance bounding box each pixel belongs to. The instance location prediction for each pixel involves the coordinates of center, top-left corner and bottom-right corner of the object instance that a specific pixel belongs to. Any off-the-self clustering method can be utilized to generate ultimate instance-level segmentation results.}}
		\label{fig:network}
		\vspace{-8mm}
	\end{center}
\end{figure*}

In this paper, we propose a simple yet effective Proposal-Free Network (PFN) for solving the instance-level segmentation task in an end-to-end way. The motivation of the proposed network is  illustrated in Figure~\ref{fig:motivation}. The pixels predicting the same instance locations can be directly clustered into the same object instance region. Moreover, the object boundaries of the occluded objects can be inferred by the difference in the predicted instance locations. For simplicity, we use the term \emph{instance locations} to denote the coordinates of the instance bounding box each pixel belongs to. Inspired by the observation that humans glance at an image and instantly know what and where the objects are in the image, we reformulate the instance-level segmentation task in the proposed network by directly inferring the regions of object instances from the global image context, in which the traditional region proposal generation step is totally disregarded. The proposed PFN framework is shown in Figure~\ref{fig:network}. To solve the semantic instance-level object segmentation task, three sub-tasks are addressed: category-level segmentation, instance location prediction for each pixel, and instance number prediction for each category in the entire image.

First, the convolutional network is fine tuned based on the pre-trained VGG classification net~\cite{vgg} to predict the category-level segmentation. In this way, the domain-specific feature representation on semantic segmentation for each pixel can be learned.

Second, by fine-tuning on the category-level segmentation network, the instance locations for each pixel as well as the instance number of each category are simultaneously predicted by the further updated instance-level segmentation network. In terms of instance locations for each pixel, six location maps including the coordinates of the center, the top left corner and the bottom right corner of the bounding box of each instance, are predicted. The predicted coordinates can be complementary to each other and make the algorithm more robust for handling close or occluded instances. To obtain more precise instance location prediction for each pixel, multi-scale prediction streams with individual supervision (i.e. multi-loss) are appended to jointly encode local details from the early, fine layers and the global semantic information from the deep, coarse layers. The feature maps from deep layers often focus on the global structure, but are insensitive to local boundaries and spatial displacement. In contrast, the feature maps from early layers can sense better the local detailed boundaries. The fusion layer combining multi-scale predictions is utilized before the final prediction layer. %The pixel-wise Euclidean distance loss is finally used to optimize the instance location predictions. During training, only the gradients within the foreground region are back-propagated for updating the network parameters.

Third, the instance numbers of all categories are described with a real number vector and also regressed with Euclidean loss in the instance-level segmentation network. Note that the instance number vector embraces the category-level information (whether the instance of a specific category appears or not) and instance-level information (how many object instances appear for a specific category). Thus, the intermediate feature maps from the category-level segmentation network and the instance-level feature maps after the fusion layer from the instance-level segmentation network are concatenated together, which can be utilized to jointly predict the instance numbers. 

In the testing stage, instance numbers and pixel-level information including category-level confidences and coordinates of the instance bounding box each pixel belongs to, can together help generate the ultimate instance-level segmentation after clustering. Note that any off-the-self clustering method can be used for this simple post-processing, and the predicted instance number specifies the exact cluster number for the corresponding category. % To retain the location sensitivity, that is, the spatially closed pixels have much higher possibility to belong to the same object instance, the coordinates of pixels are also incorporated as one part of the features for clustering. The clustering is separately performed on the segmentation mask of each category inferred from the category-level segmentation network. %To further reduce the effect of noisy background pixels within the predicted segmentation mask, the size constraints for clustered segments is also utilized. Intuitively, the predicted instance locations for noisy background pixels can be disordered, and have no connection with that of a specific object instance.

Comprehensive evaluations and comparisons on the PASCAL VOC 2012 segmentation benchmark well demonstrate that the proposed proposal-free network yields results that significantly surpass all previous published methods. It boosts the current state-of-the-art performance from 46.3\%~\cite{liu2015multi} to 58.7\%. It should be noted that all previous works utilize the extra region proposal extraction algorithms to generate the region candidates and then feed these candidates into a classification network and complex post-processing steps. Instead, our PFN generates the instance-level segmentation results in a much simple and more straightforward way.

\vspace{-3mm}
\section{Related Work}
   
Deep convolutional neural networks (DCNN) have achieved great success in object classification~\cite{szegedy2014going}~\cite{alexnet}~\cite{vgg}~\cite{wei2014cnn}, object detection~\cite{fastercnn}~\cite{redmon2015you}~\cite{babylearning}~\cite{ren2015object} and object segmentation~\cite{long2014fully}~\cite{wcrf}~\cite{BoxSup}~\cite{hariharan2014simultaneous}~\cite{noh2015learning}. In this section, we discuss the most relevant work on object detection, semantic segmentation and instance-level object segmentation.

\textbf{Object Detection.} Object detection aims to localize and recognize every object instance with a bounding box. The detection pipelines~\cite{girshick2014rich}~\cite{fastercnn}~\cite{redmon2015you}~\cite{gidaris2015object} generally start from extracting a set of box proposals from input images and then identify the objects using classifiers or localizers. The box proposals are extracted either by the hand-crafted pipelines such as selective search~\cite{uijlings2013selective}, EdgeBox~\cite{zitnick2014edge} or the designed convolutional neural network such as deep MultiBox~\cite{erhan2014scalable} or region proposal network~\cite{fastercnn}. For instance, the region proposal network~\cite{fastercnn} simultaneously predicts object bounds and objectiveness scores to generate a batch of proposals and then uses the Fast R-CNN~\cite{girshick2015fast} for detection. Different from these prior work, Redmon et al.~\cite{redmon2015you} first proposed a You Only Look Once (YOLO) pipeline that predicts bounding boxes and class probabilities directly from full images in one evaluation. Our work shares some similarities with YOLO, where the region proposal generation is discarded. However, our PFN is based on the intuition that the pixels inferring  similar instance locations can be directly collected as a single instance region. The pixel-wise instance locations and the instance number of each category are simultaneously optimized in one network. Finally, the fine-grained segmentation mask of each instance can be produced with our PFN instead of the coarse outputs depicted by the bounding boxes from YOLO.

\textbf{Semantic Segmentation.}  The most recent progress in object segmentation~\cite{long2014fully}~\cite{wcrf}~\cite{BoxSup} was achieved by fine-tuning the pre-trained classification network with the ground-truth category-level masks. For instance, Long et al.~\cite{long2014fully} proposed a fully convolutional network for pixel-wise labeling. %Dai et al.~\cite{BoxSup} estimated segmentation masks for training by extracting region proposals from the annotated boxes. 
Papandreou et al.~\cite{wcrf} utilized the foreground/background segmentation methods to generate segmentation masks, and conditional random field inference is used to refine the segmentation results. Zheng et al.~\cite{CRF-RNN} formulated the conditional random fields as recurrent neural networks for dense semantic prediction. Different from the category-level prediction by these previous methods, our PFN targets at predicting the instance-level object segmentation that provides more powerful and informative predictions to enable the real-world vision applications. Note that these previous pipelines using the pixel-wise cross-entropy loss for semantic segmentation cannot be directly utilized for instance-level segmentation because the instance number of each category for different images significantly varies, and the output size of prediction maps cannot be constrained to a pre-determined number.

\begin{figure*}
	\begin{center}
		\includegraphics[scale=0.6]{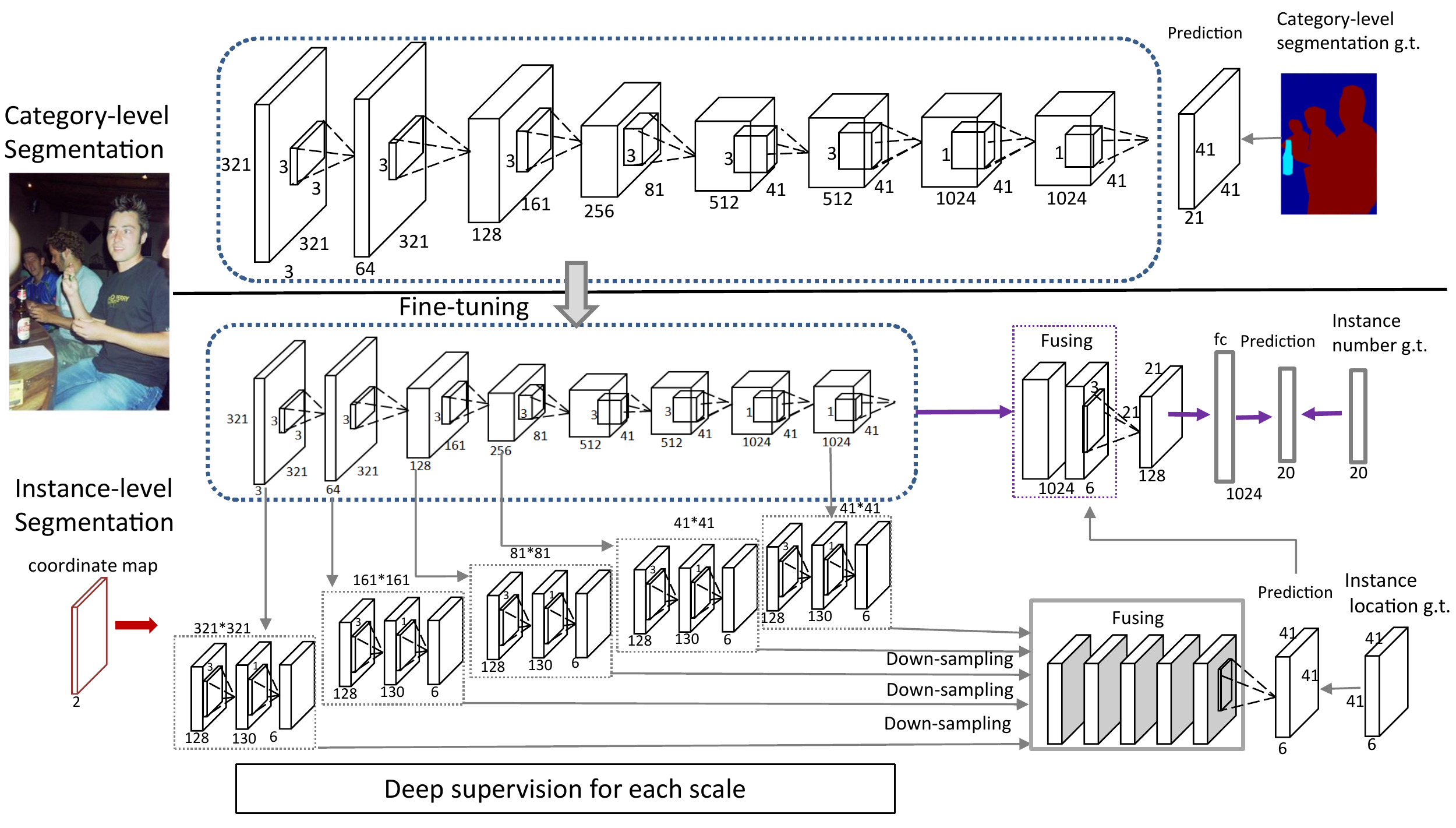}
		\vspace{-4mm}
		\caption{{The detailed network architecture and parameter setting of PFN. First, the category-level segmentation network is fine-tuned based on the pre-trained VGG-16 classification network. The cross-entropy loss is used for optimization. Second, the instance-level segmentation network that simultaneously predicts the instance numbers of all categories and the instance location vector for each pixel is further fine-tuned. The multi-scale prediction streams (with different resolution and reception fields) are appended to the intermediate convolutional layers, and are then fused to generate final instance location predictions. During each stream, we incorporate the corresponding coordinates (i.e. x and y dimension) of each pixel as the feature maps in the second convolutional layer with 130 = 128 + 2 channels. The regression loss is used during training. To predict instance numbers, the convolutional feature maps and the instance location maps are concatenated together for inference, and the Euclidean loss is used. The two losses from two targets are jointly optimized for the whole network training.}}
		\label{fig:framework}
		\vspace{-8mm}
	\end{center}
\end{figure*}

\textbf{Instance-level Object Segmentation.} Recently several approaches which tackle the instance-level object segmentation~\cite{hariharan2014simultaneous}~\cite{liu2015multi}~\cite{silberman2014instance}~\cite{zhang2015monocular}~\cite{tighe2014scene} have emerged. Most of the prior works utilize the region proposal methods as the requisite. For example, Hariharan et al.~\cite{hariharan2014simultaneous} classified region proposals using features extracted from both the bounding box and the region foreground with a jointly trained CNN. Similar to~\cite{hariharan2014simultaneous}, Chen et al.~\cite{liu2015multi} proposed to use the category-specific reasoning and shape prediction through exemplars to further refine the results after classifying the proposals.~\cite{silberman2014instance} designed a higher-order loss function to make an optimal cut in the hierarchical segmentation tree based on the region features. Other works have resorted to the object detection task to initialize the instance segmentation and the complex post-processing such as integer quadratic program~\cite{tighe2014scene} and probabilistic model~\cite{yang2012layered} to further determine the instance segments.

These prior works based on region proposals are very complicated due to several pre-processing and post-processing steps. In addition, combining independent steps is not an optimal solution because the local and global context information cannot be incorporated together for inferring. In contrast, our PFN  directly predicts pixel-wise instance location maps and uses a simple clustering technique to generate instance-level segmentation results. In particular, Zhang et al.~\cite{zhang2015monocular} predicted depth-ordered instance labels of the image patch and then combined predictions into the final labeling via the Markov Random Field inference. However, the instance number to be present in each image patch is limited to be smaller than 6 (including background), which makes it not scalable for real-world images with an arbitrary number of possible object instances. Instead, our network predicts the instance number in a totally data-driven way by the trained network, which can be naturally scalable and easily extended to other instance-level recognition tasks.

\vspace{-3mm}
\section{Proposal-Free Network}

Figure~\ref{fig:framework} shows the detailed network architecture of PFN. The category-level segmentation, instance locations for each pixel and instance numbers of all categories are taken as three targets during the PFN training. 

\vspace{-2mm}
\subsection{Category-level Segmentation Prediction}

The proposed PFN is fine-tuned based on the publicly available pre-trained VGG 16-layer classification network~\cite{vgg} for the dense category-level segmentation task. We utilize the ``DeepLab-CRF-LargeFOV" network structure as the basic presented in~\cite{DeepLabCRF} due to its leading accuracy and competitive efficiency. The important convoluational filters are shown in the top row of Figure~\ref{fig:framework}, and other intermediate convolutional layers can be found in the published model file~\cite{DeepLabCRF}. The reception field of the ``DeepLab-CRF-LargeFOV" architecture is $224\times 224$ with zero-padding, which enables effective prediction of the subsequent instance locations that requires the global image context for reasoning. For category-level segmentation, the 1000-way ImageNet classifier in the last layer of VGG-16 is replaced with $C+1$ confidence maps, where $C$ is the category number. The loss function is the sum of pixel-wise cross-entropy in terms of the confidence maps (down-sampled by 8 compared to the original image). During testing, the fully-connected conditional random fields~\cite{DeepLabCRF} 
are employed to generate more smooth and accurate segmentation maps.

This fine-tuned category-level network can generate semantic segmentation masks for the subsequent instance-level segmentation for each input image. Then the instance-level network is fine-tuned based on the category-level network, where the $C+1$ category-level predictions are eliminated. Note that we use two separate stages from optimizing category-level segmentation and instance-level segmentation. The intuition is that category-level segmentation prefers the prediction that is insensitive for different object instances of a specific category while instance-level segmentation aims to distinguish between individual instances. The motivations of two targets are significantly different. Therefore the convolutional feature maps, especially for the latter convolutional layers, cannot be shared. We verify the superiority of subsequently fine-tuning two separate networks for two tasks in the experiment. In addition, the performance on instance-level segmentation is much better when fine-tuning the instance-level network based on the category-level segmentation network compared to the original VGG-16. This may be because the category-level segmentation can provide a better start for parameter learning where the basic segmentation-aware convolutional filters have already been well learned.

\vspace{-3mm}
\subsection{Instance-level Segmentation Prediction}

The instance-level segmentation network takes an image with an arbitrary size as the input and outputs the corresponding instance locations for all the pixels and the instance number of each category.

\textbf{Pixel-wise Instance Location Prediction.} For each image, the instance location vector of each pixel is defined as the bounding box information of its corresponding object instance that contains the pixel. The object instance $s$ of a specific category can be identified by its center $(c^x, c^y)$, the top-left corner $(l^x, l^y)$ and the bottom-right corner $(r^x, r^y)$ of its surrounding bounding box, as illustrated in Figure~\ref{fig:network}. Note that the information may be redundant by using two corners besides the centers. Incorporating the redundant information can be treated as the model combination, which increases the robustness of the algorithm to noises and inaccurate prediction. For each pixel $i$ belonging to the object instance $s$, the ground-truth instance location vector is denoted as $t_{i} = (c^x_s/w_s, c^y_s/h_s, l^x_s/w_s, l^y_s/h_s, r^x_s/w_s, r^y_s/h_s)$, where $w_s$ and $h_s$ are the width and the height of the object instance $s$, respectively. With these definitions, we minimize an objective function to optimize the instance location which is inspired by the one used for Fast R-CNN~\cite{girshick2015fast}. Let $t_i$ denote the predicted location vector and $t_i^*$ the ground-truth location vector for each pixel $i$, respectively. The loss function $\ell^o$ can be defined as

\vspace{-2mm}
\begin{equation}
\ell^o(t_i, t_i^*) = [k_i^* \geq 1]R(t_i - t_i^*),
\end{equation}

\noindent{where} $k_i^* \in \{0,1,2,\dots, C\}$ is the semantic label for the pixel $i$, and $C$ is the category number. $R$ is the robust loss function (smooth-$L_1$) in~\cite{girshick2015fast}. The term $[k_i^* \geq 1]R(t_i - t_i^*)$ means the regression loss that is activated only for the foreground pixels and disabled for background pixels. The reason of using this filtered loss is that predicting the instance locations is only possible for foreground pixels which definitely belong to a specific instance.

\begin{figure*}
	\begin{center}
		\includegraphics[scale=0.55]{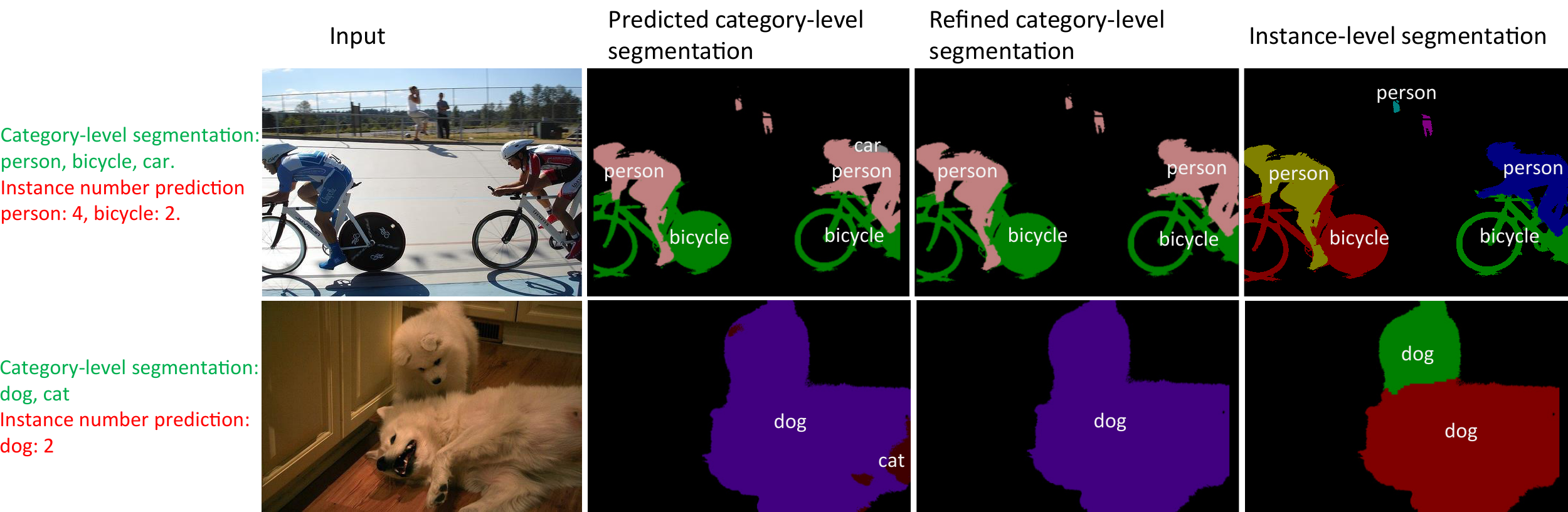}
		\vspace{-2mm}
		\caption{{The exemplar segmentation results by refining the category-level segmentation with the predicted instance numbers. For each image, we show their classification results inferred from category-level segmentation and the predicted instance numbers in the left. In the first row, the refining strategy is to convert the inconsistent predicted labels into background. In the second row, the refining strategy is to convert the wrongly predicted labels in category-level segmentation to the ones predicted in the instance number vector. Different colors indicate different object instances. Better viewed in zoomed-in color pdf file.}}
		\label{fig:testing}
		\vspace{-8mm}
	\end{center}
\end{figure*}

Following the recent results of~\cite{DeepLabCRF}~\cite{long2014fully}, we have also utilized the multi-scale prediction to increase the instance location prediction accuracy. As illustrated in Figure~\ref{fig:framework}, the five multi-scale prediction streams are attached to the input image, the output of each of the first three max pooling layers and the last convolutional layer (fc7) in the category-level segmentation network. For each stream, two layers (first layer: 128 convolutional filters, second layer: 130 convolutional filters) and deep supervision (i.e. individual loss for each stream) are utilized. The spatial padding for each convolutional layer is set so that the spatial resolution of feature maps is preserved. The multi-scale predictions from five streams are accordingly down-sampled and then concatenated to generate the fused feature maps (as the fusing layer in Figure~\ref{fig:framework}). Then the $1\times 1$ convolutional filters are used to generate the final pixel-wise predictions. It should be noted that multi-scale predictions are with different spatial resolution and inferred under different reception fields. In this way, the fine local details (e.g. boundaries and local consistency) captured by early layers with higher resolution and the high-level semantic information from subsequent layers with lower resolution can jointly contribute to the final prediction. 

To predict final instance location maps in each stream, we attach their own coordinates of all the pixels as the additional feature maps in the second convolutional layer to help generate the instance location predictions. The intuition is that predicting accurate instance locations for each pixel may be difficult due to various spatial displacements in each image while the intrinsic offsets between each pixel position and its corresponding instance locations are much easier to be learned. By incorporating the spatial coordinate maps into feature maps, more accurate location predictions can be obtained, which is also verified in our experiment. Consider that the feature maps $x^v$ of the $v$-th convolutional layer are a three-dimensional array of size $h^v \times w^v \times d^v$, where $h^v$ and $w^v$ are spatial dimensions and $d^v$ is the channel number. We generate $2$ spatial coordinate maps $[x^o_1, x^o_2]$ with size $h^v \times w^v$, where $x^o_{i_x, i_y,1}$ and $x^o_{i_x, i_y,2}$ at the spatial position $(i_x,i_y)$ for each pixel $i$ are set as $i_x$ and $i_y$, respectively. By concatenating the feature maps $x^v$ and the coordinate maps, the combined feature maps $\hat{x}^v = [x^v, x^o_1, x^o_2]$ of size $h^v \times w^v \times (d^v + 2)$ can be obtained. The outputs $x_{i_x, i_y}^{v+1}$ at the location $(i_x,i_y)$ in the next layer can be computed by

\vspace{-2mm}
\begin{equation}
x_{i_x, i_y}^{v+1} = f_b(\{\hat{x}^v_{i_x + \delta_{i_x}, i_x + \delta_{i_y}}\}_{0\leq \delta_{i_x}, \delta_{i_y}\leq b}),
\end{equation}

\noindent{where} $b$ is the kernel size and $f_b$ is the convolutional filters. In PFN, $x^{v+1}$ represents the final instance location prediction maps with six channels.

Suppose we have $M = 5$ multi-scale prediction streams, and each stream is associated with a regression loss $\ell^o_m(\cdot), m\in \{1,2,\dots,M\}$. For each image, the loss for the final prediction maps after fusing is denoted as $\ell^o_{\text{fuse}}(\cdot)$. The overall loss function for predicting pixel-wise instance locations then becomes

\vspace{-3mm}
\begin{equation}
L^o(\mathbf{t}, \mathbf{t^*}) = \sum_{m=1}^M\frac{\sum_{i}\ell^o_m(t_i, t_i^*)}{\Omega} + \frac{\sum_{i}\ell^o_{\text{fuse}}(t_i, t_i^*)}{\Omega},
\end{equation}

\noindent{where} $\mathbf{t} = \{t_i\}$ and $\mathbf{t^*} = \{t_i^*\}$ represent the predicted instance locations and ground-truth instance locations of all pixels, respectively. The $\Omega$ denotes the number of foreground pixels for each image. Divided by $\Omega$, the resulting loss $L^o$ can be prevented from being too large, which may lead to non-convergence. 

\textbf{Instance Number Prediction.} Another sub-task of PFN is to predict the instance numbers of all categories. The instance numbers of the input image that account for the object instances of each category naturally contains the category-level information and instance-level information. As shown in Figure~\ref{fig:framework}, the feature maps of the last convolutional layer from the previously trained category-level segmentation network and the instance location predictions are combined together to form the fused feature maps with $1024 + 6$ channels. These fused feature maps are then convolved with $3\times3$ convolutional filters and down-sampled with stride 2 to obtain the 128 feature maps. Then the fully-connected layer with 1024 outputs is performed to generate the final $C$-dimensional instance number prediction maps. 

Given an input image $I$, we denote the instance number vector of all $C$ categories as $\mathbf{g} = [g_1, g_2, \dots, g_C]$, where $g_c, c \in \{1,2, \dots, C\}$ represents the object instance number of each category appearing in the image. Let $\mathbf{g}$ denote the predicted instance number vector and $\mathbf{g^*}$ represent the ground-truth instance number vector for each image, respectively. The loss function $L^n$ is defined as

\vspace{-3mm}
\begin{equation}
L^n(\mathbf{g}, \mathbf{g^*}) = \frac{1}{C}\sum_{c=1}^{C} ||g_c - g^*_c||^2.
\end{equation}

\textbf{Network Training.} To train the whole instance-level network, the over loss function $L$ for each image is actuated as

\vspace{-4mm}
\begin{equation}
L(\mathbf{t}, \mathbf{t^*}, \mathbf{g}, \mathbf{g^*}) = \lambda L^o(\mathbf{t}, \mathbf{t^*}) + L^n(\mathbf{g}, \mathbf{g^*}).
\end{equation}

The class-balancing parameter $\lambda$ is empirically set to 10, which means the bias towards better pixel-wise instance location prediction. In this way, the instance number predictions and pixel-wise instance location predictions are jointly optimized in a unified network. The two different targets can benefit each other by learning more robust and discriminative shared convolutional filters. We borrow the convolutional filters except for those of the last prediction layer in the previously trained category-level network to initialize the parameters of the instance-level network. We randomly initialize all newly added layers by drawing weights from a zero-mean Gaussian distribution with standard deviation 0.01. The network can be trained by back-propagation and stochastic gradient descent (SGD)~\cite{lecun1989backpropagation}.

\begin{figure*}
	\begin{center}
		\includegraphics[scale=0.8]{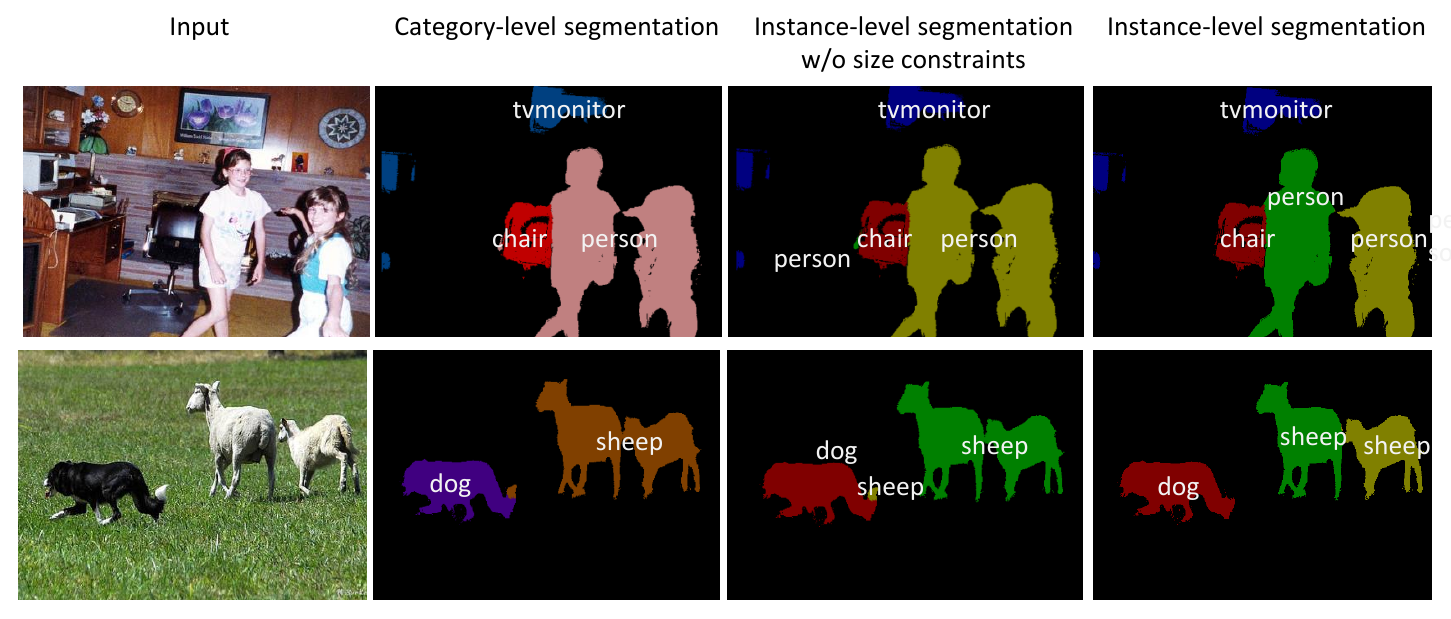}
		\vspace{-4mm}
		\caption{{Comparison of segmentation results by constraining the pixel number of each clustered object instance.  }}
		\vspace{-8mm}
		\label{fig:cluster}
	\end{center}
\end{figure*}

\vspace{-2mm}
\subsection{Testing Stage}

During testing, we first feed the input image $I$ into the category-level segmentation network to obtain category-level segmentation results, and then pass the input image into the instance-level network to get the instance number vector $\mathbf{g}$ and the pixel-wise instance location predictions $\mathbf{t}$. 

Then the clustering based on all the predicted instance locations $\mathbf{t}$ of all the pixels can be performed. We separately cluster the predicted instance locations for each category, which can be obtained by filtering the $\mathbf{t}$ with the category-level segmentation result $\mathbf{p}$, and the predicted instance numbers of all categories $\mathbf{g}$ indicate the expected cluster numbers used for spectral clustering. The simple normalized spectral clustering~\cite{ng2002spectral} is utilized due to its simplicity and effectiveness. For each category $c$, the similarity matrix $W$ is constructed by calculating the similarities between any pair of pixels that belong to the resulting segmentation mask $\mathbf{p_c}$.  Let the spatial coordinate vectors for the pixel $i$ and $j$ be $q_i = [i_x, i_y]$ and $q_j = [j_x, j_y]$, respectively. The $t_i$ and $q_i$ vectors are all normalized by their corresponding maximum. The Gaussian similarity function $w_{i,j}$ for each pair $(i,j)$ is computed as

\vspace{-3mm}
\begin{equation}
w_{i,j} = \exp(\frac{-||t_i - t_j||^2/|t_i|}{2\sigma^2}) + \exp(\frac{-||q_i - q_j||^2/|q_i|}{2\sigma^2}),
\label{eq:sim}
\end{equation}
where $|t_i|$ denotes the feature dimension for the vector $t_i$, which equals 6 (the coordinates of centers, top-left corner and bottom-right corner), and $|q_i|$ indicates the feature dimension of $q_i$, which equals 2. During clustering, we simply connect all pixels in the same segmentation mask of a specific category with positive similarity because the local neighboring relationships can be captured by the second term in Eqn.~(\ref{eq:sim}). We simply set $\sigma = 0.5$ for all images. To make the clustering results robust to the initialization of seeds during the $\emph{k}$-means step of spectral clustering, we randomly select the seeds twenty times by balancing the accuracy and computational cost. Then the clustering result with maximal average within-cluster similarities for all clusters is selected as the final result.

Note that inconsistent global image category predictions from instance number vectors and pixel-wise category-level segmentation are often observed. For example, as illustrated in the first row of Figure~\ref{fig:testing}, the instance number prediction infers 4 person instances and 2 bicycle instances while the category-level segmentation indicates three categories (i.e. person, bicycle, car) appearing in the image. Thus it is necessary to keep the predicted global image category to be consistent between the instance number prediction and the pixel-wise segmentation. Note that the instance number prediction task is much simpler than pixel-wise semantic segmentation due to dense pixel-wise optimization targets. We can thus use instance number prediction to refine the produced category-level segmentation. 

The object category from instance number prediction can be easily obtained by thresholding the instance number vector by $\tau = 0.5$, which means, if the predicted instance number of a specific category $c$ is larger than $\tau$, the category $c$ is regarded as the true label. Specifically, two strategies are adopted: first, if more than one category is predicted to have at least one instance in the image, any pixels assigned with all other categories (i.e. the categories with predicted instance number 0) will be re-labeled as the background, as illustrated in the first row of Figure~\ref{fig:testing}; second, if only one category is inferred from instance number prediction, pixels labeled with other object categories (excluding background pixels) in the semantic segmentation mask will be totally converted into the predicted ones, as illustrated in the second row of Figure~\ref{fig:testing}. The refined category-level segmentation masks are used to further generate instance-level segments.
 
In addition, the predicted segmentation result is not perfect due to the noisy background pixels. The instance locations of pixels belonging to one object have much higher possibilities to form a cluster while the predictions of background pixels are often quite random, forming very small clusters. Therefore, we experimentally discard those clusters, whose pixel numbers are less than 0.1\% of the pixels in the segmentation mask. Finally, after obtaining the final clustering result for each category, the instance-level object segmentation result can be easily obtained by combining all the clustering results of all categories. Example results after constraining the pixel number of each clustered instance region are shown in Figure~\ref{fig:cluster}.

\section{Experiments}

\subsection{Experimental Settings}
\textbf{Dataset and Evaluation Metrics.} The proposed PFN is extensively evaluated on the PASCAL VOC 2012 validation segmentation benchmark~\cite{everingham2014pascal}. We compare our method with two state-of-the-art algorithms: SDS~\cite{hariharan2014simultaneous} and the method by Chen et al.~\cite{liu2015multi}. Following two baselines~\cite{hariharan2014simultaneous}~\cite{liu2015multi}, the segmentation annotations from SBD~\cite{hariharan2011semantic} are used for training the network, {and 1,449 images in the PASCAL VOC 2012 segmentation validation set are used for evaluation. We cannot report results on PASCAL VOC 2012 segmentation test set because no instance-level segmentation annotation is provided. In addition, because VOC 2010 segmentation set is only a subset of VOC 2012 and no baseline has reported results on VOC 2010, we only evaluate our algorithm on VOC 2012 set.} For fair comparison with state-of-the-art instance level segmentation methods, $AP^r$ and $AP^r_{vol}$ metrics are used following SDS~\cite{hariharan2014simultaneous}. The $AP^r$ metric measures the average precision under 0.5 IoU overlap with ground-truth segmentation. Hariharan et al.~\cite{liu2015multi} proposed to vary IoU scores from 0.1 to 0.9 to show the performance for different applications. The $AP^r_{vol}$ metric calculates the mean of $AP^r$ under all IoU scores. Note that two baselines fine-tune the networks based on Alexnet architecture~\cite{krizhevsky2012imagenet}. For fair comparison, we also report results based on the Alexnet architecture~\cite{krizhevsky2012imagenet}. %Note that our PFN can only generate at most one segment for each object instance due to the clustering strategy. In this way, the results of our method equal to the top detection results under the settings of the state-of-the-art methods~\cite{hariharan2014simultaneous}~\cite{liu2015multi} that use several region proposals for one instance.

%To evaluate the performances of our instance number prediction, we use the standard Mean Absolute Error (MAE)~\cite{geng2007automatic} as the evaluation metric, which is widely used for the regression problem, such as face age estimation. The MAE can be calculated by:
%
%\begin{equation}
%MAE_c = \frac{1}{N}\sum_{n=1}^{N} |g_{n,c} - g^*_{n,c}|,
%\label{eq:sim}
%\end{equation}
%
%where $g_{n,c}$ is the predicted instance number and $g^*_{n,c}$ is the groundtruth instance number for the $n$-th image on $c$-th category.

\textbf{Training Strategies.} All networks in our experiments are trained and tested based on the published DeepLab code~\cite{wcrf}, which is implemented based on the publicly available Caffe platform~\cite{jia2014caffe} on a single NVIDIA GeForce Titan GPU with 6GB memory. We first train the category-level segmentation network, which is then used to initialize our instance-level segmentation network for fine tuning. For both training stages, we randomly initialize all new layers by drawing weights from a zero-mean Gaussian distribution with standard deviation 0.01. We use mini-batch size of 8 images, initial learning rate of 0.001 for pre-trained layers, and 0.01 for newly added layers in all our experiments. We decrease the learning rate to 1/10 of the previous one after 20 epochs and train the two networks for roughly 60 epochs one after the other. The momentum and the weight decay are set as 0.9 and 0.0005, respectively. The same training setting is utilized for all our compared network variants.

We evaluate the testing time by averaging the running time for images on the VOC 2012 validation set on NVIDIA GeForce Titan GPU and Intel Core i7-4930K CPU @3.40GHZ. Our PFN can rapidly process one $300\times 500$ image in about one second. This compares much favorably to other state-of-the-art approaches, as the current state-of-the-art methods~\cite{hariharan2014simultaneous}~\cite{liu2015multi} rely on region proposal pre-processing  and complex post-processing steps: ~\cite{hariharan2014simultaneous} takes about 40 seconds while ~\cite{liu2015multi} is excepted to be more expensive than ~\cite{hariharan2014simultaneous} because more complex top-down category specific reasoning and shape prediction are further employed for inference based on ~\cite{hariharan2014simultaneous}.

\begin{table*} \setlength{\tabcolsep}{0.7pt}
	\centering
	
	\caption{Comparison of instance-level segmentation performance with several architectural variants of our network and two state-of-the-arts using $AP^r$ metric over 20 classes at 0.5 IoU on the PASCAL VOC 2012 validation set. All numbers are in \%. }\renewcommand\arraystretch{1.7}\vspace{-3mm}
	\begin{tabular}{l|c|cccccccccccccccccccc|c }
		\toprule    
		Settings & Method &\rotatebox{90}{plane}&\rotatebox{90}{bike}&\rotatebox{90}{bird}&\rotatebox{90}{boat}&\rotatebox{90}{bottle}&\rotatebox{90}{bus}&\rotatebox{90}{car}&\rotatebox{90}{cat}&\rotatebox{90}{chair}&\rotatebox{90}{cow}&\rotatebox{90}{table}&\rotatebox{90}{dog}&\rotatebox{90}{horse}&\rotatebox{90}{motor}&\rotatebox{90}{person}&\rotatebox{90}{plant}&\rotatebox{90}{sheep}&\rotatebox{90}{sofa}&\rotatebox{90}{train}&\rotatebox{90}{tv}& average\\
		\midrule
		\hline
		\multirow{2}*{Baselines} & SDS~\cite{hariharan2011semantic} & 58.8 & 0.5 & 60.1 & 34.4 & 29.5 & 60.6 & 40.0 & 73.6 & 6.5 & 52.4 & 31.7 & 62.0 & 49.1 & 45.6 & 47.9 & 22.6 & 43.5 & 26.9 & 66.2 & 66.1 & 43.8\\
		& Chen et al.~\cite{liu2015multi} & 63.6 & 0.3 & 61.5 & 43.9 & 33.8 & 67.3 & 46.9 & 74.4 & 8.6 & 52.3 & 31.3 & 63.5 & 48.8 & 47.9 & 48.3 & 26.3 & 40.1 & 33.5 & 66.7 & 67.8 & 46.3\\
		\hline
		\multirow{1}*{Ours (Alexnet)} & PFN Alexnet & 63.0 & 15.3 & 69.8 & 48.4 & 23.5 & 60.2 & 24.1 & 82.2 & 13.9 & 60.7 & 41.3 & 73.5 & 76.9 & 69.7 & 37.6 & 20.0 & 41.4 & 58.8 & 78.9 & 58.4 & 50.9\\
		\hline
		\multirow{1}*{Training Strategy} & PFN unified & 72.9 & \textbf{18.1} & 78.8 & 55.4 & 23.2 & 63.6 & 17.8 & 72.1 & 14.7 & 64.1 & 44.5 & 69.5 & 71.5 & 63.3 & 39.1 & 9.5 & 27.9 & 47.7 & 72.1 & 57.0 & 49.1\\
		\hline
		\multirow{5}*{Location prediction} & PFN offsets(2)& 73.6 & 17.2 & 78.9 & 55.6 & \textbf{29.4} & 61.6 & 31.7 & 77.8 & 13.5 & 59.6 & 38.4 & 70.2 & 67.5 & 66.8 & 43.6 & 9.8 & 41.1 & 43.3 & 75.3 & 65.2 & 51.0\\
		& PFN centers(2)& 78.0 & 15.5 & 76.4 & 58.7 & 25.6 & 69.3 & 28.9 & 88.2 & 16.6 & 67.2 & 47.8 & 82.3 & 78.0 & 71.5 & 47.0 & 23.8 & 48.8 & 63.8 & 83.3 & 72.0 & 57.1\\
		& PFN centers,w,h(4)& 80.1 & 15.9 & 76.6 & 60.2 & 25.7 & 70.9 & 30.0 & 87.7 & 18.3 & 70.1 & 50.8 & 82.5 & 77.9 & 71.4 & 47.4 & 24.4 & 48.0 & 64.0 & 82.8 & 72.2 & 57.8\\
		& PFN centers,topleft(4)& \textbf{80.5} & 15.9 & \textbf{79.2} & \textbf{62.5} & 27.8 & 69.4 & 31.1 & 86.6 & \textbf{18.8} & 73.6 & 50.6 & 81.6 & 77.2 & 71.5 & 47.5 & 22.5 & 47.7 & 63.6 & 83.0 & \textbf{72.5} & 58.2\\
		& PFN +topright, bottomleft(10)& 76.9 & 15.1 & 73.9 & 55.8 & 26.0 & 73.7 & 31.1 & 92.1 & 17.6 & 74.0 & 48.1 & 82.0 & \textbf{85.5} & 71.7 & \textbf{48.8} & \textbf{25.2} & \textbf{57.7} & 64.5 & 88.7 & 72.0 & \textbf{59.0}\\
		\hline
		\multirow{3}*{Network structure}& PFN w/o multiscale& 72.8 & 16.5 & 71.9 & 50.3 & 25.2 & 65.9 & 27.4 & 90.4 & 16.3 & 64.6 & 48.1 & 78.9 & 74.1 & 72.0 & 42.4 & 21.3 & 46.2 & 64.4 & 82.8 & 72.3 & 55.2\\
		& PFN w/o coordinate maps& 74.1 & 17.3 & 72.8 & 57.0 & 27.6 & 72.8 & 30.0 & \textbf{92.6} & 17.7 & 69.4 & 48.5 & 81.7 & 80.9 & 72.0 & 47.8 & 24.9 & 50.1 & 62.7 & 87.9 & 72.3 & 58.0\\
		& PFN fusing\_summation& 80.6 & 16.5 & 78.8 & 59.4 & 27.2 & 71.0 & 29.9 & 85.1 & 18.1 & \textbf{75.4} & \textbf{52.8} & 80.9 & 76.5 & \textbf{74.1} & 47.7 & 20.8 & 46.7 & \textbf{66.8} & 86.0 & 71.7 & 58.3\\
		\hline
		\multirow{3}*{Instance number} & PFN w/o category-level& 72.4 & 17.6 & 77.7 & 55.4 & \textbf{29.4} & 63.5 & \textbf{32.0} & 77.8 & 13.3 & 61.2 & 38.5 & 70.8 & 69.5 & 66.1 & 44.3 & 13.1 & 42.6 & 45.6 & 71.9 & 64.5 & 51.4\\
		& PFN w/o instance-level& 74.0 & 15.6 & 72.9 & 55.5 & 26.4 & 72.2 & 31.3 & 91.1 & 19.1 & 66.9 & 49.9 & 82.3 & 75.0 & 73.4 & 47.7 & 24.2 & 53.3 & 64.4 & \textbf{91.0} & 72.3 & 57.9\\
		& PFN separate\_finetune& 74.1 & 17.3 & 74.3 & 57.2 & 27.6 & 72.8 & 30.2 & \textbf{92.6} & 18.1 & 69.7 & 49.5 & 81.7 & 80.9 & 71.6 & 47.8 & 24.9 & 50.1 & 62.7 & 87.9 & 71.8 & 58.1\\
		\hline
		\multirow{4}*{Testing strategy} 
		& PFN w/o coordinates& 72.7 & 16.8 & 72.0 & 51.8 & 24.0 & 67.6 & 27.4 & 90.4 & 16.7 & 64.6 & 48.1 & 78.9 & 74.1 & 71.5 & 42.8 & 22.1 & 45.2 & 64.4 & 82.8 & 72.3 & 55.3\\
		& PFN w/o classify+size& 76.2 & 16.1 & 72.9 & 57.9 & 25.3 & 69.5 & 29.4 & 88.3 & 15.8 & 67.1 & 48.1 & 82.0 & 73.8 & 71.8 & 46.1 & 22.3 & 47.8 & 62.7 & 83.7 & 72.3 & 56.4\\
		& PFN w/o size& 72.0 & 17.3 & 72.8 & 57.0 & 27.6 & 72.8 & 30.8 & \textbf{92.6} & 17.7 & 64.6 & 48.5 & 81.7 & 80.9 & 73.3 & \textbf{48.8} & 24.9 & 50.1 & 62.7 & 87.9 & 72.3 & 57.8\\
		& PFN w/o classify& 71.0 & 15.6 & 72.9 & 55.3 & 25.8 & 70.4 & 31.4 & 91.1 & 17.7 & 66.9 & 48.5 & \textbf{82.7} & 76.4 & 72.3 & 47.2 & 24.2 & 51.4 & 64.4 & \textbf{91.0} & 72.3 & 57.4\\
		
		\hline
		Ours (VGG 16) & PFN & 76.4 & 15.6 & 74.2 & 54.1 & 26.3 & \textbf{73.8} & 31.4 & 92.1 & 17.4 & 73.7 & {48.1} & 82.2 & 81.7 & 72.0 & 48.4 & 23.7 & \textbf{57.7} & 64.4 & 88.9 & 72.3 & 58.7\\
		\hline
		\multirow{3}*{Upperbound}& PFN upperbound\_instnum& 81.6 & 19.0 & 80.0 & 58.1 & 30.0 & 77.0 & 33.9 & 92.9 & 19.8 & 82.6 & 57.2 & 81.3 & 83.0 & 74.4 & 49.6 & 21.6 & 56.2 & 67.8 & 91.2 & 68.9 & 61.3\\
		& PFN upperbound\_instloc & 81.8 & 23.6 & 84.6 & 66.4 & 38.2 & 75.3 & 35.3 & 94.9 & 24.8 & 84.2 & 61.7 & 83.9 & 87.2 & 75.2 & 55.6 & 27.3 & 63.9 & 69.3 & 88.9 & 72.5 & 64.7\\
		\midrule
	\end{tabular}
	\label{mpr}
	\vspace{-8mm}
\end{table*}

\subsection{Results and Comparisons}

In the first training stage, we train a category-level segmentation network using the same architecture as ``DeepLab-CRF-LargeFOV" in~\cite{wcrf}. By evaluating the pixel-wise segmentation in terms of pixel intersection-over-union (IOU)~\cite{long2014fully} averaged across 21 classes, we archive 67.53\% on category-level segmentation task on the PASCAL VOC 2012 validation set~\cite{everingham2014pascal}, which is only slightly inferior to $67.64\%$ reported in~\cite{wcrf}. %Note that our network mainly focuses on fine-grained instance-level segmentation task, we just utilized the public semantic segmentation network as previous work~\cite{wcrf} for simplicity and easy implementation.

Table~\ref{mpr} and Table~\ref{mpr_vol} present the comparison of the proposed PFN with two state-of-the-art methods~\cite{hariharan2011semantic}~\cite{liu2015multi} using $AP^r$ metric at  IoU score 0.5 and 0.6 to 0.9, respectively. We directly use their published results on PASCAL VOC 2012 validation set for fair comparison. All results of the state-of-the-art methods were reported in~\cite{liu2015multi} which re-evaluated the performance of~\cite{hariharan2011semantic} using VOC 2012 validation set. For fair comparison, we also report the results of PFN using the Alexnet architecture~\cite{krizhevsky2012imagenet} as used in two baselines~\cite{hariharan2011semantic}~\cite{liu2015multi}, i.e. ``PFN Alexnet". Following the strategy presented in~\cite{wcrf}, we convert the fully connected layers in Alexnet to fully convolutional layers, and all other settings are the same as those used in ``PFN".  The results of~\cite{hariharan2011semantic} and~\cite{liu2015multi} achieve $43.8\%$ and $46.3\%$ in $AP^r$ metric at IoU  0.5. Meanwhile, our ``PFN Alexnet" is significantly superior over the two baselines, i.e. 50.9\% vs 43.8\%~\cite{hariharan2011semantic} and 46.3\%~\cite{liu2015multi} in $AP^r$ metric. Further detailed comparisons in $AP^{r}$ over 20 classes at  IoU scores 0.6 to 0.9 are listed in Table~\ref{mpr_vol}. By utilizing the more powerful VGG-16 network architecture, our PFN can substantially improve the performance and outperform these two baselines by over $14.9\%$ for SDS~\cite{hariharan2011semantic} and $12.4\%$ for Chen et al.~\cite{liu2015multi}. PFN also gives huge boosts in $AP^r$ metrics at 0.6 to 0.9 IoU scores, as reported in Table~\ref{mpr_vol}. For example, when evaluating at 0.9 IoU score where the localization accuracy for object instances is strictly required, the two baselines achieve $0.9\%$ for SDS~\cite{hariharan2011semantic} and 2.6\% for ~\cite{liu2015multi} while PFN obtains $15.7\%$. This verifies the effectiveness of our PFN although it does not require extra region proposal extractions as the pre-processing step. The detailed $AP^r$ scores for each class are also listed. In general, our method shows dramatically higher performance than the baselines. Especially, in predicting small object instances (e.g., bird and chair) or object instances with a lot of occlusion (e.g., table and sofa), our method achieves a larger gain, e.g. 74.2\% vs 60.1\%~\cite{hariharan2011semantic} and 61.5\%~\cite{liu2015multi} for bird, 64.4\% vs 26.9\%~\cite{hariharan2011semantic} and 33.5\%~\cite{liu2015multi} for sofa. This demonstrates that our network can effectively deal with the internal boundaries between the object instances and robustly predict the instance-level masks with various appearance patterns or occlusion. In Table~\ref{compare_mpr_vol}, we also report the $AP^r_{vol}$ results of our different architecture variants, which average all $AP^r$ at 0.1 to 0.9 IoU scores. We cannot compare the $AP^r_{vol}$ with the baselines as they~\cite{liu2015multi} do not publish these results. 

\begin{table*} \setlength{\tabcolsep}{0.7pt}
	\centering
	
	\caption{Comparison of instance-level segmentation performance with several architectural variants of our network using $AP^r_{vol}$ metric over 20 classes that averages all $AP^r$ performance from 0.1 to 0.9 IoU scores on the PASCAL VOC 2012 validation set. All numbers are in \%.}\renewcommand\arraystretch{1.7}\vspace{-3mm}
	\begin{tabular}{l|c|cccccccccccccccccccc|c }
		\toprule    
		Settings & Method &\rotatebox{90}{plane}&\rotatebox{90}{bike}&\rotatebox{90}{bird}&\rotatebox{90}{boat}&\rotatebox{90}{bottle}&\rotatebox{90}{bus}&\rotatebox{90}{car}&\rotatebox{90}{cat}&\rotatebox{90}{chair}&\rotatebox{90}{cow}&\rotatebox{90}{table}&\rotatebox{90}{dog}&\rotatebox{90}{horse}&\rotatebox{90}{motor}&\rotatebox{90}{person}&\rotatebox{90}{plant}&\rotatebox{90}{sheep}&\rotatebox{90}{sofa}&\rotatebox{90}{train}&\rotatebox{90}{tv}& average\\
		\hline
		\multirow{1}*{Ours (Alexnet)} & PFN Alexnet & 62.2 & 20.0 & 64.6 & 41.0 & 23.8 & 56.6 & 22.4 & 76.0 & 15.5 & 56.2 & 39.2 & 68.6 & 65.4 & 61.2 & 35.2 & 20.2 & 37.6 & 52.1 & 68.9 & 49.9 & 46.8\\
		\hline
		\multirow{1}*{Training Strategy} & PFN unified & 71.2 & 22.8 & \textbf{74.1} & 47.3 & 24.2 & 55.1 & 18.5 & 69.8 & 15.4 & 56.2 & 40.1 & 63.7 & 63.0 & 56.2 & 38.1 & 13.2 & 31.5 & 41.6 & 63.8 & 47.1 & 45.6\\
		\hline
		\multirow{5}*{Location prediction} & PFN offsets(2)& 70.4 & \textbf{23.1} & 73.4 & 46.6 & 31.0 & 55.3 & 29.1 & 74.3 & 15.6 & 54.7 & 35.4 & 64.9 & 60.0 & 57.7 & 41.3 & 13.5& 42.6 & 39.6 & 64.7 & 53.2 & 47.3\\
		& PFN centers(2)& 72.5 & 21.0 & 69.6 & 50.0 & 25.0 & 63.3 & 27.2 & 79.2 & 17.1 & 60.8 & 44.3 & 74.9 & 67.8 & 64.3 & 41.0 & 24.3 & 42.5 & 55.6 & 72.8 & 58.0 & 51.6\\
		& PFN centers,w,h(4)& \textbf{74.4} & 21.8 & 69.8 & 51.3 & 25.1 & 64.7 & 28.2 & 78.6 & 18.4 & 63.3 & \textbf{46.9} & 74.0 & 67.6 & 64.3 & 41.2 & \textbf{24.7} & 41.6 & 55.4 & 72.6 & 58.4 & 52.1\\
		& PFN centers,topleft(4)& 74.3 & 21.8 & 72.8 & \textbf{53.0} & 27.3 & 63.4 & 29.0 & 77.7 & 18.6 & \textbf{65.7} & 46.8 & 73.3 & 67.1 & 65.0 & 41.5 & 23.2 & 40.7 & 54.7 & 72.8 & 58.7 & 52.4\\
		& PFN +topright, bottomleft(10)& 71.3 & 20.8 & 66.4 & 48.5 & 26.6 & 65.2 & 27.2 & 83.1 & 17.3 & 64.6 & 45.1 & 74.6 & 70.8 & 64.3 & 41.6 & 23.4 & \textbf{48.8} & 56.4 & 76.1 & 57.9 & \textbf{52.5}\\
		\hline
		\multirow{3}*{Network structure}& PFN w/o multiscale& 69.6 & 21.3 & 66.7 & 44.1 & 25.8 & 58.4 & 25.3 & 81.9 & 17.4 & 60.6 & 44.5 & 72.8 & 64.7 & 62.4 & 39.5 & 21.6 & 41.6 & \textbf{57.7} & 74.2 & 58.7 & 50.4\\
		& PFN w/o coordinate maps& 70.5 & 21.9 & 65.2 & 50.0 & 26.6 & \textbf{67.6} & 27.8 & 83.1 & 17.9 & 61.4 & 45.1 & 73.9 & \textbf{68.8} & 62.8 & 41.9 & 24.1 & 44.8 & 56.4 & 76.5 & 58.7 & 52.3\\
		& PFN fusing\_summation& 74.3 & 22.7 & 72.4 & 50.3 & 27.0 & 64.6 & 27.7 & 76.2 & 17.7 & 66.4 & 48.6 & 72.7 & 66.6 & \textbf{67.1} & 41.7 & 21.7 & 39.8 & 57.0 & 75.3 & 57.8 & 52.4\\
		\hline
		\multirow{3}*{Instance number} & PFN w/o category-level& 69.4 & 22.3 & 71.9 & 46.5 & \textbf{31.5} & 58.0 & \textbf{29.5} & 74.1 & 15.8 & 56.9 & 35.4 & 66.1 & 61.8 & 57.8 & 42.1 & 15.8 & 43.3 & 40.9 & 63.6 & 52.6 & 47.8\\
		& PFN w/o instance-level& 69.4 & 21.1 & 65.1 & 48.6 & 25.6 & 64.3 & 27.4 & 81.9 & \textbf{18.7} & 61.1 & 46.0 & \textbf{75.2} & 67.0 & 64.0 & 41.0 & 22.4 & 47.6 & 57.6 & \textbf{77.7} & 58.8 & 52.0\\
		& PFN separate\_finetune& 70.5 & 21.9 & 66.3 & 50.2 & 26.6 & \textbf{67.6} & 27.9 & 83.1 & 18.3 & 61.7 & 45.8 & 73.9 & \textbf{68.8} & 62.4 & 42.0 & 24.1 & 44.8 & 56.4 & 76.5 & 58.1 & 52.3\\
		\hline
		\multirow{4}*{Testing strategy} 
		& PFN w/o coordinates& 69.9 & 21.7 & 66.7 & 44.4 & 25.3 & 58.9 & 25.3 & 81.9 & 17.9 & 60.6 & 44.5 & 72.8 & 64.7 & 63.3 & 39.5 & 22.1 & 41.9 & \textbf{57.7} & 74.2 & 58.7 & 50.6\\
		& PFN w/o classify+size& 70.7 & 21.3 & 66.6 & 49.4 & 24.6 & 63.4 & 27.7 & 79.3 & 16.3 & 60.8 & 44.5 & 74.6 & 64.7 & 64.6 & 40.2 & 22.7 & 41.7 & 54.9 & 73.2 & 58.4 & 51.0\\
		& PFN w/o size& 68.8 & 22.0 & 65.3 & 50.0 & 26.6 & 67.0 & 28.5 & 83.1 & 17.9 & 59.3 & 45.1 & 74.3 & \textbf{68.8} & 64.6 & \textbf{42.9} & 23.7 & 44.9 & 56.4 & 77.5 & 58.4 & 52.3\\
		& PFN w/o classify& 67.8 & 21.1 & 65.3 & 48.3 & 25.7 & 62.3 & 27.6 & 81.9 & 17.7 & 60.0 & 44.9 & \textbf{75.2} & 67.0 & 62.9 & 41.1 & 22.8 & 45.0 & 57.6 & \textbf{77.7} & \textbf{58.8} & 51.5\\
		
		\hline
		Ours (VGG 16) & PFN & 70.8 & 21.1 & 66.7 & 47.6 & 26.7 & 65.3 & 27.5 & \textbf{83.2} & 17.2 & {64.5} & 45.1 & 74.7 & 67.9 & 64.5 & 41.3 & 22.1 & \textbf{48.8} & 56.5 & 76.2 & 58.2 & 52.3\\
		\midrule
	\end{tabular}
	\label{compare_mpr_vol}
	\vspace{-3mm}
\end{table*}

\vspace{-3mm}
\subsection{Ablations Studies of Our networks}
We further evaluate the effectiveness of our six important components of PFN, including the training strategy, instance location prediction, network structure, instance number prediction, testing strategy and upperbounds, respectively. The performance over all the categories by all variants is reported in Table~\ref{mpr} and Table~\ref{compare_mpr_vol}.

\textbf{Training strategy:} Note that our PFN training includes two stages: the category-level segmentation network and the instance-level network. To justify the necessity of using two stages, we evaluate the performance of training a unified network that consists of the category-level segmentation, pixel-wise instance location prediction and instance number prediction in one learning stage, namely ``PFN unified". ``PFN unified" is fine-tuned based on the VGG-16 pre-trained model and three losses for three sub-tasks are optimized in one network. The category-level prediction is appended in the last convolutional layer within the dashed blue box in Figure~\ref{fig:framework}, and the loss weight for category-level segmentation is set as 1. From our experimental results, ``PFN unified" leads to 9.6\% decrease in average $AP^r$ and $6.6\%$ decrease in average $AP^r_{vol}$, compared with ``PFN". Intuitively, the target of category-level segmentation is to be robust for individual object instances of the same category while erasing the instance-level information during optimization. On the contrary, the instance-level network aims to learn the instance-level information for distinguishing different object instances with large variance in appearance, view or scale. This comparison result verifies well that training with two sequential stages can lead to better global instance-level segmentation.

\begin{table*} \setlength{\tabcolsep}{2pt}
	\centering
	
	\caption{Per-class instance-level segmentation results using $AP^r$ metric over 20 classes at 0.6 to 0.9 (with a step size of 0.1) IoU scores on the VOC PASCAL 2012 validation set. All numbers are in \%.}\renewcommand\arraystretch{1.7}\vspace{-3mm}
	\begin{tabular}{l|c|cccccccccccccccccccc|c }
		\toprule    
		IoU score & Method &\rotatebox{90}{plane}&\rotatebox{90}{bike}&\rotatebox{90}{bird}&\rotatebox{90}{boat}&\rotatebox{90}{bottle}&\rotatebox{90}{bus}&\rotatebox{90}{car}&\rotatebox{90}{cat}&\rotatebox{90}{chair}&\rotatebox{90}{cow}&\rotatebox{90}{table}&\rotatebox{90}{dog}&\rotatebox{90}{horse}&\rotatebox{90}{motor}&\rotatebox{90}{person}&\rotatebox{90}{plant}&\rotatebox{90}{sheep}&\rotatebox{90}{sofa}&\rotatebox{90}{train}&\rotatebox{90}{tv}& average\\
		\midrule
		\hline
		\multirow{3}*{0.6} & SDS~\cite{hariharan2011semantic} & 43.6 & 0 & 52.8 & 19.5 & 25.7 & 53.2 & 33.1 & 58.1 & 3.7 & 43.8 & 29.8 & 43.5 & 30.7 & 29.3 & 31.8 & 17.5 & 31.4 & 21.2 & 57.7 & \textbf{62.7} & 34.5\\
		& Chen et al.~\cite{liu2015multi} & 57.1 & 0.1 & 52.7 & 24.9 & \textbf{27.8} & 62.0 & \textbf{36.0} & 66.8 & 6.4 & 45.5 & 23.3 & 55.3 & 33.8 & 35.8 & 35.6 &  \textbf{20.1} &  35.2 & 28.3 & 59.0 & 57.6 & 38.2\\
		& PFN Alexnet& 60.3 & 9.9 & 67.4 & 31.8 & 18.7 & 52.9 & 18.6 & 75.6 & 8.1 & 54.6 & 36.0 & 71.3 & 63.3 & 65.6 & 29.3 & 14.8 & 31.2 & 48.5 & 66.9 & 47.3 & 43.6\\
		& PFN & \textbf{73.2} &  \textbf{11.0}  & \textbf{70.9} & \textbf{41.3} & 22.2 & \textbf{66.7} & 26.0 & \textbf{83.4} & \textbf{10.7} & \textbf{65.0} & \textbf{42.4} & \textbf{78.0} & \textbf{69.2} & \textbf{72.0} & \textbf{38.0} & 19.0 & \textbf{46.0} & \textbf{51.8} & \textbf{77.9} & 61.4 & \textbf{51.3}\\
		\midrule
		\hline
		\multirow{3}*{0.7} & SDS~\cite{hariharan2011semantic} & 17.8 & 0 & 32.5 & 7.2 & 19.2 & 47.7 & 22.8 & 42.3 & 1.7 & 18.9 & 16.9 & 20.6 & 14.4 & 12.0 & 15.7 & 5.0 & 23.7 & 15.2 & 40.5 & 51.4 & 21.3\\
		& Chen et al.~\cite{liu2015multi} & 40.8 & 0.07 & 40.1 & 16.2 & \textbf{19.6} & 56.2 & \textbf{26.5} & 46.1 & 2.6 & 25.2 & 16.4 & 36.0 & 22.1 & 20.0 & 22.6 & 7.7 & 27.5 & 19.5 & 47.7 &  46.7 & 27.0\\
		& PFN Alexnet& 56.1 &  5.0 & 59.8 & 25.6 & 12.7 & 50.4 & 15.5 & 69.3 & 3.2 & 42.9 & 24.5 & 63.6 & 58.4 & 54.4 & 21.1 & 7.9 & 26.2 & 39.9 & 59.1 & 37.0 & 36.6\\
		& PFN & \textbf{68.5} & \textbf{5.6}  & \textbf{60.4} & \textbf{34.8} & 14.9 & \textbf{61.4} & 19.2 & \textbf{78.6} & \textbf{4.2} & \textbf{51.1} & \textbf{28.2} & \textbf{69.6} & \textbf{60.7} & \textbf{60.5} & \textbf{26.5} & \textbf{9.8} & \textbf{35.1} & \textbf{43.9} & \textbf{71.2} & \textbf{45.6} & \textbf{42.5}\\
		\midrule
		\hline
		\multirow{3}*{0.8} & SDS~\cite{hariharan2011semantic} & 2.1 & 0 & 8.3 & 4.5 & 11.5 & 32.3 & 9.0 & 17.9 & 0.7 & 4.7 & 9.0 & 6.5 & 1.8 & 4.4 & 3.3 & 1.9 & 7.9 & 10.2 & 12.7 & 24.3 & 8.7\\
		& Chen et al.~\cite{liu2015multi} & 10.5 & 0 & 15.7 & 9.8 & \textbf{11.4} & 32.7 & 12.5 & 34.8 & 1.1 & 11.6 & 9.5 & 15.3 & 4.6 & 6.5 & 6.0 & 3.0 & 13.9 & 14.4 & 27.0 & 30.4 & 13.5\\
		& PFN Alexnet& 46.6 & 1.5 & 48.0 & 15.0 & 8.6 & 44.3 & 10.7 & 58.6 & 2.0 & 37.4 & 15.5 & 47.8 & 40.0 & 34.9 & 13.5 & 3.8 & 20.2 & 23.4 & 51.7 & 30.7 & 27.7\\
		& PFN & \textbf{54.6} & \textbf{1.5}  & \textbf{49.5} & \textbf{21.0} & 10.4 & \textbf{50.7} & \textbf{14.2} & \textbf{63.5} & \textbf{2.1} & \textbf{38.3} & \textbf{18.9} & \textbf{51.8} & \textbf{41.2} & \textbf{36.7} & \textbf{16.5} & \textbf{4.2} & \textbf{26.2} & \textbf{25.3} & \textbf{59.6} & \textbf{36.9} & \textbf{31.2}\\
		\midrule
		\hline
		\multirow{3}*{0.9} & SDS~\cite{hariharan2011semantic} & 0 & 0 & 0.2 & 0.3 & 2.0 & 3.8 & 0.2 & 0.9 & 0.1 & 0.2 & 1.5 & 0 & 0 & 0 & 0.1 & 0.1 & 0 & 2.3 & 0.2 & 5.8 & 0.9\\
		& Chen et al.~\cite{liu2015multi} & 0.6 &  0  & 0.6 & 0.5 & \textbf{4.9} & 9.8 & 1.1 & 8.3 & 0.1 & 1.1 & 1.2 & 1.7 & 0.3 & 0.8 & 0.6 & 0.3 & 0.8 & 7.6 & 4.3 & 6.2 & 2.6\\
		& PFN Alexnet& 37.1 &  0.1 & 24.6 & 7.0 & 3.6 & 30.4 & 4.9 & 40.0 & \textbf{0.6} & 23.3 & 2.8 & 28.9 & \textbf{13.6} & 7.8 & 5.0 & 1.1 & 10.9 & 12.2 & 23.8 & 8.1 & 14.3\\
		& PFN & \textbf{43.9} & \textbf{0.1}  & \textbf{24.5} & \textbf{7.8} & 4.1 & \textbf{32.5} & \textbf{6.3} & \textbf{42.0} & \textbf{0.6} & \textbf{25.7} & \textbf{3.2} & \textbf{31.8} & {13.4} & \textbf{8.1} & \textbf{5.9} & \textbf{1.6} & \textbf{14.8} & \textbf{14.3} & \textbf{25.0} & \textbf{8.5} & \textbf{15.7}\\
		\midrule
		
	\end{tabular}
	\label{mpr_vol}
	\vspace{-4mm}
\end{table*}

\textbf{Instance location prediction:} Recall that PFN predicts the spatial coordinates (6 dimensions) of the center, top-left corner and bottom-right corner of the bounding box for each pixel. We also extensively evaluate other five kinds of instance location predictions: 1) ``PFN offsets (2)", which predicts the offsets of each pixel with respect to the centers of its object instance; 2) ``PFN centers (2)", where only the coordinates of its object instance center for each pixel are predicted; 3) ``PFN centers, w,h(4)", which predicts the centers, width and height of each instance; 4) ``PFN centers, topleft(4)", where the centers and top-left corners of each instance are predicted; 5) ``PFN +topright, bottomleft (10)", which additionally predicts the top-right corners and bottom-left corners of each instance besides the ones in ``PFN". The performance is obtained by changing the channel number in the final prediction layer accordingly.

The ``PFN offsets (2)" gives dramatically inferior performance to ``PFN" (51.0\% vs 58.7\% in $AP^r$ and 47.3\% vs 52.3\% in $AP^r_{vol}$). The main reason may be that offsets are possibly with negative values, which may be difficult to optimize. By only predicting the centers of an object instance, the resulting $AP^r$ of ``PFN centers (2)" also shows inferior performance to ``PFN". The reason for this inferiority may be that predicting the top-left corners and bottom-right corners can bring more information about object scales and spatial layouts. ``PFN centers, topleft(4)" and ``PFN centers, w,h(4)" theoretically predict the same information about instance locations, and also achieve similar results in $AP^r$, 57.8\% vs 58.2\%, respectively. It is worth noting that the 6 dimension predictions of ``PFN" capture redundant information compared to ``PFN centers, topleft(4)" and ``PFN centers, w,h(4)". The superiority of ``PFN" over ``PFN centers, topleft(4)" and ``PFN centers, w,h(4)" (0.5\% and 0.9\%, respectively) can be mainly attributed to the effectiveness of model combination. In our primary experiment, the version that predicts the top-left and  bottom-right corner coordinates achieves unnoticeable performance with ``PFN centers, topleft(4)", because the information captured by centers and top-left corners are equal to that by top-left and bottom-right corners. The combined features used for clustering with additional information can be equally regarded as multiple model combination, which is widely used for the object classification challenge~\cite{szegedy2014going}. Moreover, we test the performance of introducing two additional points (top-right corner and bottom-left corner) as the prediction. Only a slight improvement (0.3\%) by comparing ``PFN +topright, bottomleft (10)" with ``PFN" can be observed yet more parameters and computation memory are required. We thus only adopt the setting of predicting the center, top-left and bottom-right corners for all our other experiments.

\textbf{Network Structure:} Extensive evaluations on different network structures are also performed. First, the effectiveness of multi-scale prediction is verified. ``PFN w/o multiscale" shows the performance of using only one straightforward layer to predict pixel-wise instance locations. The performance decreases by $3.5\%$ in $AP^r$ compared with ``PFN". This significant inferiority demonstrates the effectiveness of multi-scale fusing that incorporates the local fine details and global semantic information into predicting the pixel-wise instance locations. 

Note that the spatial coordinates of each pixel are utilized as the feature maps for predicting the instance locations. The superiority of using spatial coordinates can be demonstrated by comparing ``PFN w/o coordinate maps" with ``PFN", i.e. 0.7\% difference in $AP^r$. The coordinate maps can help the feature maps be more precise for predicting absolute spatial layouts of object instances, where the convolutional filters can put more focus on learning the relative spatial offsets. 

In the fusing layer for predicting pixel-wise instance locations, ``PFN" utilizes the concatenation operation instead of element-wise summation for multi-scale prediction. ``PFN fusing\_summation" shows 0.4\% decrease in $AP^r$ when compared to ``PFN". The $1\times1$ convolutional filters are utilized to adaptively weigh the contribution of the instance location prediction of each scale, which is more reasonable and experimentally effective than simple summation.
   
\textbf{Instance Number Prediction:} We explore other options to predict the instance numbers of all categories for each image. ``PFN w/o category-level" only utilizes the instance location predictions as the feature maps for predicting instance numbers and the category-level information is totally ignored. The large gap between ``PFN w/o category-level" and ``PFN" (51.4\% vs 58.7\%) verifies the importance of using category-level information for predicting instance numbers. Because the instance lcoation predictions only capture the instance numbers of all categories and category-level information is discarded, the exact instance number for a specific category thus cannot be inferred. The importance of incorporating instance-level information is also verified by comparing ``PFN w/o instance-level" with ``PFN", 57.9\% vs 58.7\% in $AP^r$. This shows that the instance number prediction can benefit from the pixel-wise instance location prediction, where more fine-grained annotations (pixel-wise instance-level locations) are provided for learning better feature maps.

We also evaluate the performance of sequentially optimizing the instance locations and the instance number instead of using one unified network. ``PFN separate\_finetune" first optimizes the network for predicting pixel-wise instance locations, and then fixes the current network parameters and only trains the newly added parameters for instance number prediction. The performance decrease of ``PFN separate\_finetune" compared to ``PFN" (58.1\% vs 58.7\% in $AP^r$) shows well the effectiveness of training one unified network. The information in the global aspect from instance numbers can be utilized for predicting more accurate instance locations.

\textbf{Testing Strategy:} We also test different strategies for generating final instance-level segmentations during testing. Note that during spectral clustering, the similarity of two pixels is computed by considering both the prediction instance locations with 6 dimensions and two spatial coordinates of each pixel. By eliminating the coordinates in the similarity function, a significant decrease in $AP^r$ can be observed by comparing ``PFN w/o coordinates" with ``PFN", 55.3\% vs 58.7\%. This verifies that the spatial coordinates can enhance the local neighboring connections during clustering, which can lead to more reasonable and meaningful instance-level segmentation results.

Recall that two steps are used for post-processing, including refining the segmentation results with instance number prediction and constraining the cluster size during clustering. We extensively evaluate the effectiveness of using these two steps. By comparing $56.4\%$ of ``PFN w/o classify + size" that eliminates these two steps with $58.7\%$ of ``PFN" in $AP^r$, it can be observed that better performance can be obtained by leveraging the instance number prediction and constraining the cluster size to refine instance-level segmentation. After only eliminating the refining strategy by constraining the cluster size, $0.9\%$ decrease can be observed. It demonstrates that constraining the cluster size can help reduce the effect of noisy background pixels to some extent and more robust instance level segmentation results can be obtained. On the other hand, the incorporation of instance number prediction can help improve the performance in $AP^r$ by $1.3\%$ when comparing ``PFN w/o classify" with ``PFN". In particular, significant improvements can be observed for easily confused categories such as 73.7\% vs 66.9\% for ``cow", 57.7\% vs 51.4\% for ``sheep" and 81.7\% vs 76.4\% for ``horse". This demonstrates the effectiveness of using instance number prediction for refining the pixel-wise segmentation results.

\textbf{Upperbound:} Finally, we also evaluate the limitations of our current algorithm. First, ``PFN upperbound\_instnum" reports the performance of using the ground-truth instance number prediction for clustering and other experimental settings are kept the same. It can be seen that only $2.6\%$ improvement in $AP^r$ is obtained. The errors from instance number prediction are already small and have only little effect on the final instance-level segmentation. Second, the upperbound for instance location predictions is reported in ``PFN upperbound\_instloc" by using the ground-truth instance locations for each pixel as the features for clustering. The large gap between $64.7\%$ of ``PFN upperbound\_instloc" and $58.7\%$  of ``PFN" verifies that the accurate instance location prediction is critical for good instance-level segmentation. Note that the current category-level segmentation only achieves $67.53\%$ of pixel-wise IoU score, which largely limits the performance of our instance-level segmentation because we perform the clustering  on the category-level segmentation. A better category-level segmentation network architecture can definitely help improve the performance of instance-level segmentation under our PFN framework. 

\vspace{-4mm}
\subsection{Visual Illustration}

\begin{figure*}
	\begin{center}
		\includegraphics[scale=1.0]{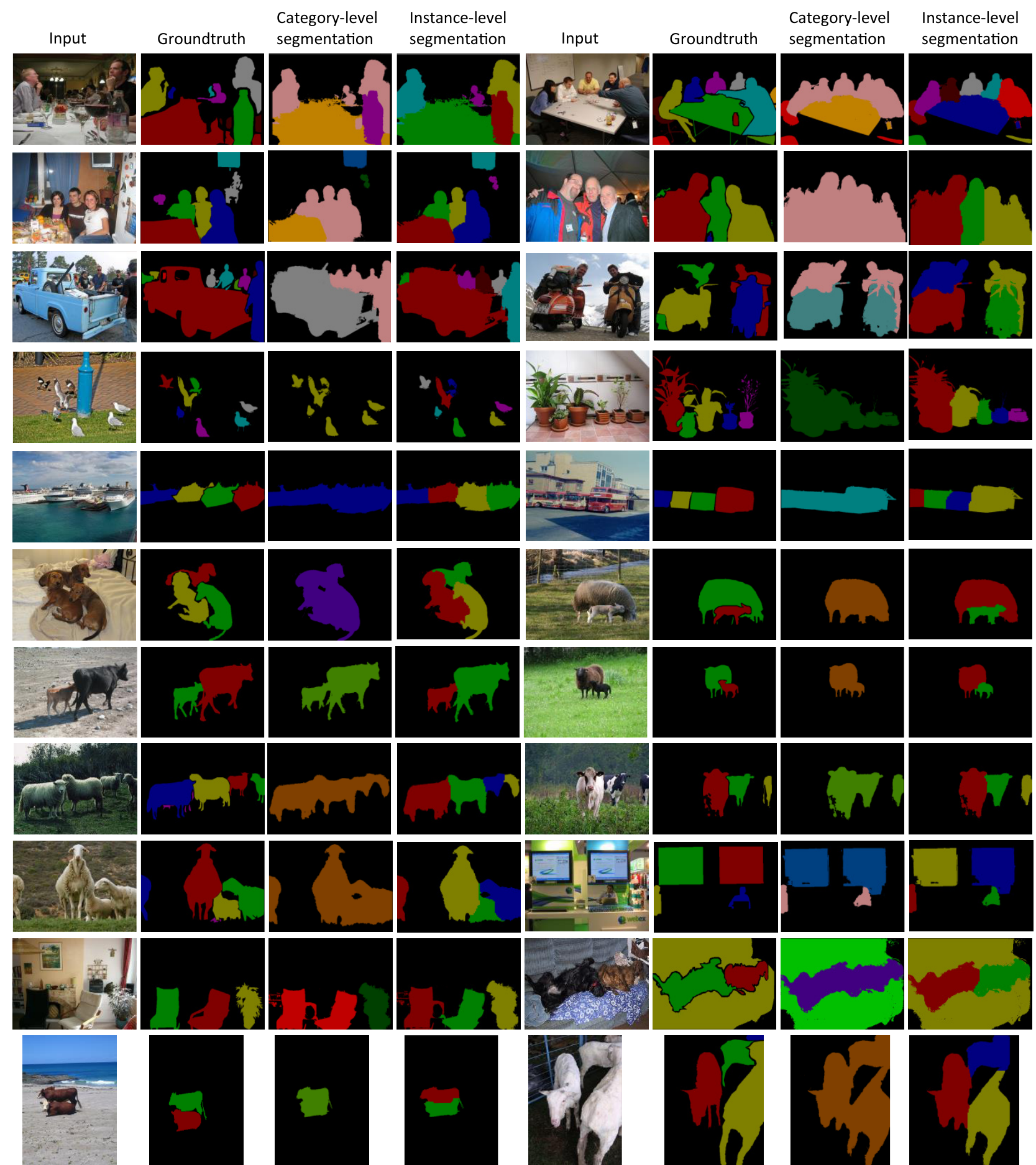}
		\caption{{Illustration of instance-level object segmentation results by the proposed PFN. For each image, we show the ground-truth instance-level segmentation, the category-level segmentation and the predicted instance-level segmentation results sequentially. Note that for instance-level segmentation results, different colors only indicate different object instances and do not represent the semantic categories. In terms of category-level segmentation, different colors are used to denote different semantic labels. Best viewed in color.}}
		\label{fig:visual}
	\end{center}
\end{figure*}

Figure~\ref{fig:visual} visualizes the predicted instance-level segmentation results with our PFN. Note that we cannot visually compare with two state-of-the-art methods because they generate several region proposals for each instance and can only visualize top detection results for each image as described in their papers. However, our method can directly produce exact region segments for each object instance just like the results of category-level segmentation. Different colors indicate different object instances for the instance-level segmentation results. The semantic labels of our instance-level segmentation results are exactly the same with the ones labeled in category-level segmentation results. It can be observed that the proposed PFN performs well in predicting the object instances with heavy occlusion, large background clutters and complex scenes. For example, the first three rows show the results on images with very complex background clutters, several objects with heavy occlusion and diverse appearance patterns. The predicted instance-level segmentations are highly consistent with the ground-truth annotations, and the object instances with heavy occlusion can also be visually distinguished, such as the first images in the first and third rows. The fourth row illustrates some images with very small object instances, such as birds and potted-plants. The fifth row shows examples where the object instances in one image have high similarity in appearance and much occlusion with each other. Other results show more examples of instance-level images under diverse scenarios and with very challenging poses, scales, views and occlusion. These visualization results further demonstrate the effectiveness of the proposed PFN.

\begin{figure}
	\begin{center}
		\includegraphics[scale=0.35]{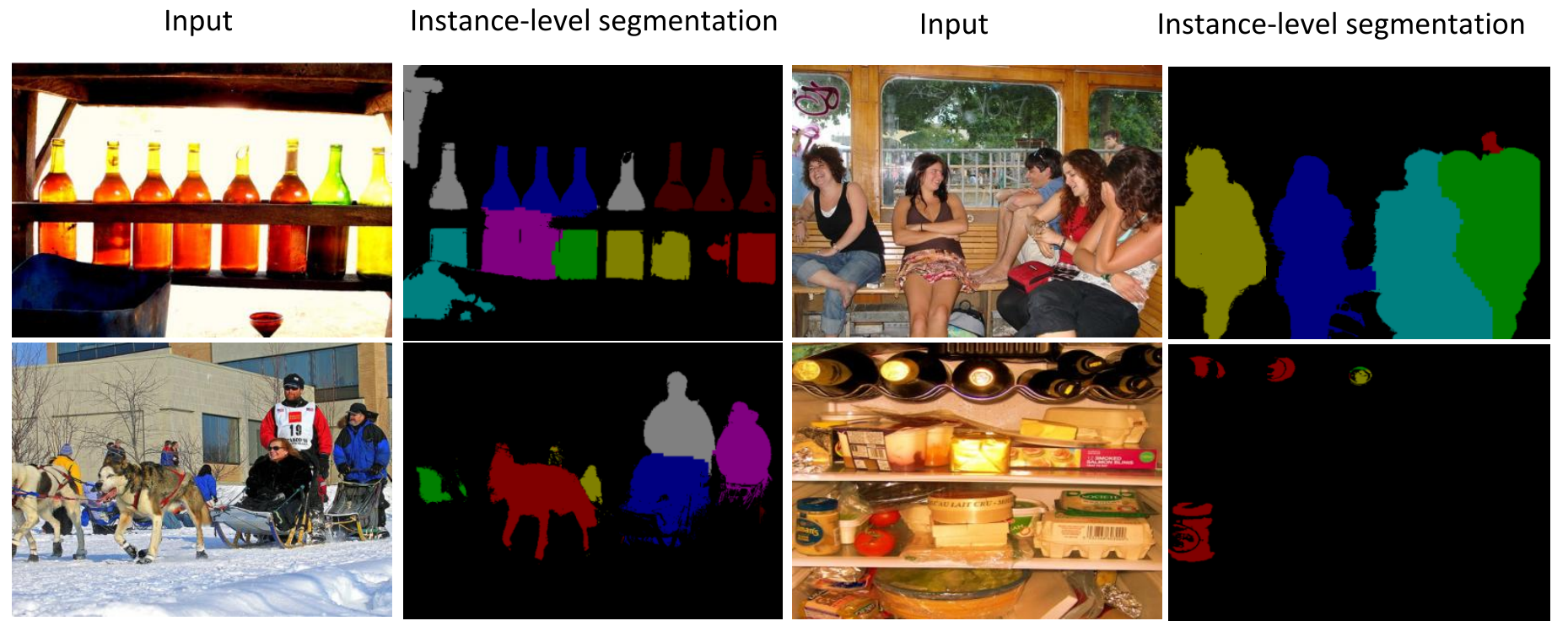}
			\vspace{-3mm}
		\caption{{Illustration of failure cases. Our PFN may fail to segment object instances with heavy occlusion (in first row) and small instances (in second row).}}
		\vspace{-8mm}
		\label{fig:failure}
	\end{center}
\end{figure}

{We also show some failure cases of our PFN in Table~\ref{fig:failure}. The instances with heavy occlusion and some small object instances are difficult to identify and segment. In future, we can make more efforts to tackle these more challenging instances.} 

\section{ Conclusion and Future Work}
{In this paper, we present an effective proposal-free network for fine-grained instance-level segmentation. Instead of utilizing extra region proposal methods as the requisite, PFN directly predicts the instance location vector for each pixel that belongs to a specific instance and the instance numbers of all categories. The pixels that predict the same or close instance locations can be directly regarded as belonging to the same object instance. During testing, the simple spectral clustering is performed on the predicted pixel-wise instance locations to generate the final instance-level segmentation results, and the predicted instance numbers of all categories are employed to indicate the cluster number for each category. Significant improvements over the state-of-the-art methods are achieved by PFN on the PASCAL VOC 2012 segmentation benchmark. Extensive evaluations of different components of PFN are conducted to validate the effectiveness of our method. Our PFN, without complicated per-processing and post-processing as requisite, is much simpler to implement with much lower computational cost compared with previous state-of-the-arts. In the future, we plan to extend our framework to the generic multiple instances in outdoor and indoor scenes, which may have higher degrees of clutters and occlusion.  

\vspace{-2mm}
\bibliographystyle{ieee}
\bibliography{proposalfree_ver4}

\begin{IEEEbiography}[{\includegraphics[width=1in,height=1.25in,clip,keepaspectratio]{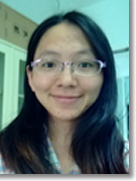}}]{Xiaodan Liang} is a Ph.D. student from Sun Yat-sen University, China. She is currently working at National University of Singapore as a Research Intern. Her research interests mainly include semantic segmentation, object/action recognition and medical image analysis.
\end{IEEEbiography}

\begin{IEEEbiography}[{\includegraphics[width=1in,height=1.25in,clip,keepaspectratio]{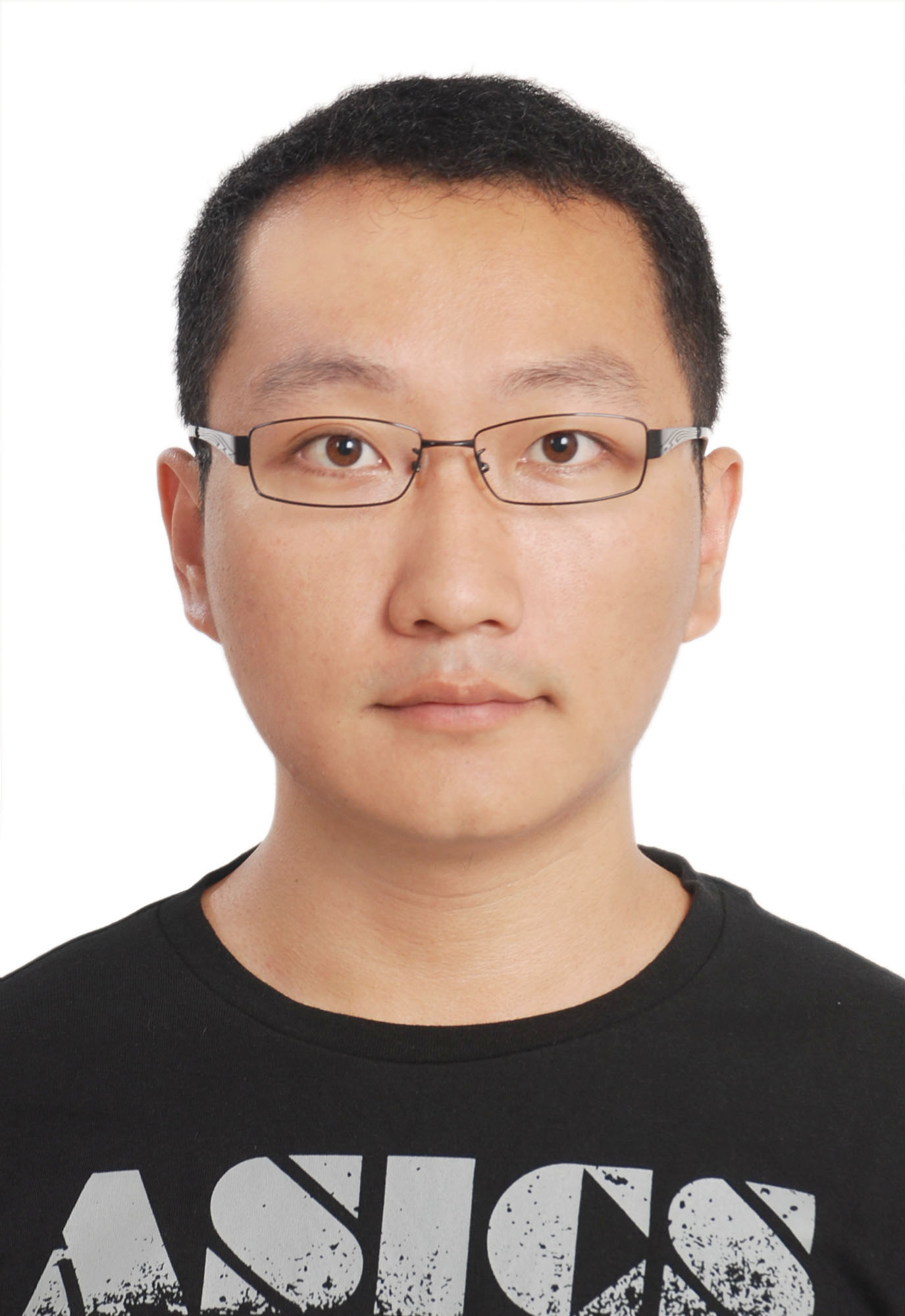}}]{Yunchao Wei} is a Ph.D. student from the Institute of Information Science, Beijing Jiaotong University, China. He is currently working at National University of Singapore as a Research Intern. His research interests mainly include object classification in computer vision and multi-modal analysis in multimedia.
\end{IEEEbiography}

\begin{IEEEbiography}[{\includegraphics[width=1in,height=1.25in,clip,keepaspectratio]{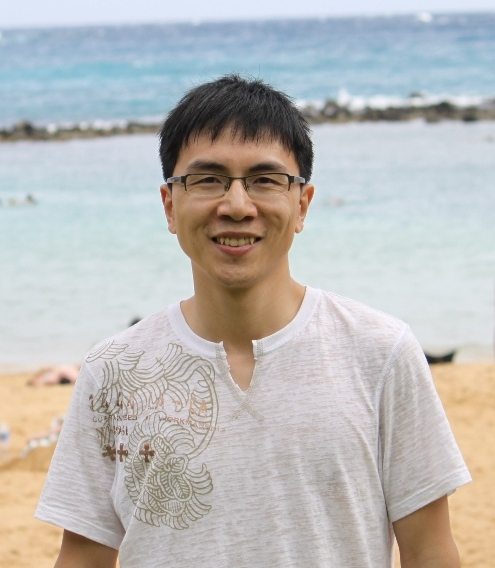}}]{Xiaohui Shen} received his PhD degree from the Department of EECS at Northwestern University in 2013. Before that, he received the MS and BS degrees from the Department of Automation at Tsinghua University, China. He is currently a research scientist at Adobe Research, San Jose, CA. His research
	interests include image/video processing and computer vision.
\end{IEEEbiography}

\begin{IEEEbiography}[{\includegraphics[width=1in,height=1.25in,clip,keepaspectratio]{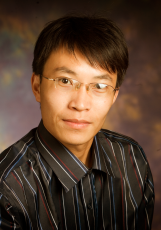}}]{Jianchao Yang}(S’08, M’12) received the M.S. and Ph.D. degrees in electrical and computer engineering from the University of Illinois at Urbana-Champaign, Urbana, in 2011. His research interests include object recognition, deep learning, sparse coding, image/video enhancement, and deblurring.  
\end{IEEEbiography}

\begin{IEEEbiography}[{\includegraphics[width=1in,height=1.25in,clip,keepaspectratio]{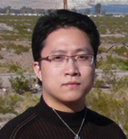}}]{Liang Lin} is a full Professor with the School of Advanced Computing, Sun Yat-Sen University (SYSU), China. He received the Ph.D. degree from the Beijing Institute of Technology (BIT), Beijing, China, in 2008. He was a Post-Doctoral Research Fellow with the Center for Vision, Cognition, Learning, and Art of UCLA. His research focuses on new models, algorithms and systems for intelligent processing and understanding of visual data such as images and videos. He was supported by several promotive programs or funds for his works, such as “Program for New Century Excellent Talents” of Ministry of Education (China) in 2012, and Guangdong NSFs for Distinguished Young Scholars in 2013. He received the Best Paper Runners-Up Award in ACM NPAR 2010, Google Faculty Award in 2012, and Best Student Paper Award in IEEE ICME 2014. 
\end{IEEEbiography}

\begin{IEEEbiography}[{\includegraphics[width=1in,height=1.25in,clip,keepaspectratio]{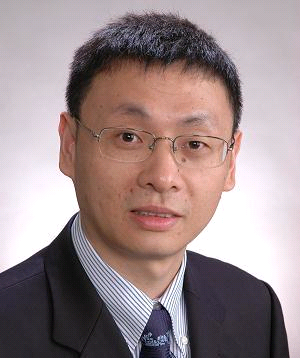}}]{Shuicheng Yan} (M'06-SM'09) is currently an Associate Professor at the Department of Electrical and Computer Engineering at National University of Singapore, and the founding lead of the Learning and Vision Research Group (http://www.lv-nus.org). Dr. Yan's research areas include machine learning, computer vision and multimedia, and he has authored/co-authored {nearly 400} technical papers over a wide range of research topics, with Google Scholar {citation$>$12,000} times. He is ISI highly-cited researcher 2014, and IAPR Fellow 2014. He has been serving as an associate editor of IEEE TKDE, CVIU and TCSVT. He received the Best Paper Awards from ACM MM'13 (Best Paper and Best Student Paper), ACM MM’12 (Best Demo), PCM'11, ACM MM’10, ICME’10 and ICIMCS'09, the runner-up prize of ILSVRC'13, the winner prizes of the classification task in PASCAL VOC 2010-2012, the winner prize of the segmentation task in PASCAL VOC 2012, the honorable mention prize of the detection task in PASCAL VOC'10, 2010 TCSVT Best Associate Editor (BAE) Award, 2010 Young Faculty Research Award, 2011 Singapore Young Scientist Award, and 2012 NUS Young Researcher Award.
\end{IEEEbiography}
\end{document}